%% file: paper.tex
\documentclass{elsarticle}

\usepackage{amssymb}

\usepackage{graphicx}
\usepackage{amsmath}
\usepackage{algorithmic}
\usepackage{algorithm}
\usepackage{subfig}
\usepackage{url}
\usepackage{times}
\usepackage{multirow}

\newtheorem{Lemma}{Lemma}
\newtheorem{Coro}{Corollary}

\def\sign{\mathop{\rm sign}}
\input{labvision_defs}

\journal{Pattern Recognition}

\begin{document}

\begin{frontmatter}

\title{BAdaCost: Multi-class Boosting with Costs}

\author[afb]{Antonio Fern\'andez-Baldera}
\ead{antonio.fbaldera@upm.es}

\cortext[cor1]{Corresponding author}
\author[jmb]{Jos\'e M. Buenaposada\corref{cor1}}
\ead{josemiguel.buenaposada@urjc.es}

\author[lb]{Luis Baumela}
\ead{lbaumela@fi.upm.es}

\address[afb,lb]{Universidad Polit\'ecnica de Madrid,
ETSI Inform{\'a}ticos.\\
Campus Montegancedo s/n, 
28660 Boadilla del Monte, Spain}

\address[jmb]{Universidad Rey Juan Carlos, ETSII.
C/ Tulip\'an, s/n, 28933 M\'ostoles, Spain}

\begin{abstract}
  We present BAdaCost, a multi-class cost-sensitive classification
  algorithm. It combines a set of cost-sensitive multi-class weak
  learners to obtain a strong classification rule within the Boosting
  framework.
  To derive the algorithm we introduce CMEL, a 
  \emph{Cost-sensitive Multi-class Exponential Loss} that generalizes
  the losses optimized in various classification
  algorithms such as AdaBoost, SAMME, Cost-sensitive AdaBoost and
  PIBoost. Hence unifying them under a common theoretical framework.
  In the experiments performed we prove that BAdaCost achieves
  significant gains in performance when compared to previous multi-class
  cost-sensitive approaches. The advantages of the proposed algorithm
  in asymmetric multi-class classification are also evaluated
  in practical multi-view face and car detection problems.
\end{abstract}

\begin{keyword}
Boosting \sep Multi-class classification \sep Cost-sensitive classification \sep Multi-view object detection
\end{keyword}

\end{frontmatter}


\section{Introduction}\label{Sec:Intro}

Boosting algorithms have been extensively used in many
computer vision problems, such as 
facial expression~\cite{Ali16} and
gait~\cite{Ma17}
recognition, 
 but particularly in object and face
detection~\cite{Viola04,Mathias14,Ohn-Bar15,Ren16}.
They learn a ``strong'' classifier by iteratively combining simple or
``weak'' predictors. Their popularity is based on their simplicity,
good generalization, feature selection capability and fast
performance when used in a cascade
framework~\cite{Viola04,Mathias14,Zhang07,Dollar14}.  Boosting
algorithms are conceived to minimize the number of incorrect
predictions. However, it is well known that this is a bad strategy
when solving asymmetric problems such as those with large data
imbalances~\cite{Sun06,SunY07,Tang17,Sheng17}, or those with very different
class priors or costs~\cite{Elkan01,Masnadi07,Masnadi11}.

Various cost-sensitive Boosting
approaches have been proposed with the aim of addressing asymmetric problems.
Most of them consider the standard two-class
case~\cite{Masnadi11,Ting00,Viola04,Viola01,Landesa12}.  They
introduce the asymmetry in the Boosting algorithm by tuning the
classifier decision threshold~\cite{Viola04} or manipulating the
weight update process~\cite{Ting00,Viola01}.  These
modifications, however, are not guaranteed to converge to a good
classification rule.  Masnadi and Vasconcelos introduce a binary
cost-sensitive algorithm with asymptotic convergence to the optimal
classification rule~\cite{Masnadi11}. This is the first theoretically
sound cost-sensitive Boosting algorithm in the binary classification
context. However, it is not applicable in asymmetric multi-class
situations such as, for example, those arising when detecting in images
several objects or a single object in different configurations.

Very few works in the Boosting literature address the multi-class
cost-sensitive situation.  The first attempt was AdaC2.M1~\cite{Sun06}
that combined the re-weighting process developed for AdaC2~\cite{SunY05}
with the AdaBoost.M1 schedule \cite{Freund97}.  This simple heuristic
algorithm has the same problem than its multi-class predecessor when a
negative predictor confidence 
emerges because it is ``too weak.'' 
Another proposal resorted to the minimization of a $p$-norm based cost
functional~\cite{Lozano08}.
However any $L_p$-CSB variant requires a relational hypothesis (weak
learners that multiply the size of data by the number of labels) which
is an important drawback with large datasets.
The MultiBoost algorithm \cite{Wang13} is based on the minimization of a
new cost-sensitive multi-class loss function.
However, it does not generalize any
previous approaches and requires an imprecise pool of multi-class weak
learners to work.

In this paper we introduce a well founded multi-class cost-sensitive
Boosting algorithm, BAdaCost, that stands for \emph{Boosting Adapted
  for Cost matrix}.
To this end we introduce a multi-class cost-sensitive margin that relates
multi-class margins with the costs of the decisions and
derive the  algorithm as a stage-wise minimization
of the expected exponential loss.
We prove that this algorithm is a generalization of previous
multi-class~\cite{Zhu09,Baldera14} and binary
cost-sensitive~\cite{Masnadi11} Boosting algorithms.

We validate the algorithm by comparing it with previous multi-class
cost sensitive approaches on several data sets from
the UCI repository. The experiments show that BAdaCost achieves a
significant improvement in performance. Moreover, we also evaluate our
algorithm in multi-view detection settings, that pose a highly
asymmetrical multi-class classification problem. In our experiments we
show that BAdaCost has several advantages in this context when
compared to the usual one-vs-background approach that uses one
detector per class:
\begin{itemize}
\item We slide a $K$-classes multi-class detector across the
  image instead of $K$ binary detectors. 
  Since the multi-class detector shares features between
  classes~\cite{Torralba04} it can work with
  considerably less decision trees nodes, see
  section~\ref{Subsec:Efficiency}.
\item Costs may be used to modify pair-wise class boundaries. 
  We can reduce the number of errors between positive
  classes (e.g. different orientations) and improve detection
  rates when object classes have different aspect ratios, see
  section~\ref{Subsec:CarDetect}.
\item To speed up detection our algorithm is amenable for
  cascade calibration~\cite{Zhang07}.
\end{itemize}

A preliminary version of BAdaCost appeared in~\cite{Baldera2015} for
solving imbalanced classification problems.  Here we present it in
the more general context of cost-sensitive classification, provide the
proofs that support the algorithm and relate it with other approaches
in the literature, and apply it to address the multi-view object
detection problem.

The rest of the paper is organized as
follows. Sections~\ref{Sec:Boosting} and~\ref{Sec:CostBoosting} review
the literature on multi-class and cost-sensitive Boosting.  In Section
\ref{Sec:Costs} we introduce the new multi-class cost-sensitive margin
and the BAdaCost algorithm together with various theoretical results
relating it with previous approaches in the literature.  Finally, in
Sections~\ref{Sec:Experiments} and~\ref{Sec:Conclusions} we
experimentally validate the algorithm and draw conclusions.  We
provide proofs and additional information as supplementary material.


\section{Boosting}
\label{Sec:Boosting}

In this section we briefly review some Boosting results directly
related to our proposal.
First we introduce some notation.  We use the terms \emph{label} and
\emph{class} interchangeably. The set of labels is
$L = \{ 1,2,\ldots,K \}$. The domain of the classification problem is
$X$.  Instances are $(\vx, l)$, with $\vx \in X$ and $l \in L$. We
consider $N$ training data instances $\left\{ (\vx_n ,l_n) \right\}$.
Besides, $I( \cdot )$ is the indicator function ($1$ when
argument is true, $0$ when false).  $M$ is the number of Boosting
iterations and $\vw = \{ \omega(n) | n=1,\ldots,N \}$ is a weighting
vector. We use capital letters, e.g. $T$ or $H$, for denoting
classifiers whose co-domain is a finite set of integer numbers, like
$L$. Classifiers having a set of vectors as co-domain are represented with
small bold letters, e.g. $\vf$ or $\vg$.


\subsection{Binary Boosting}

AdaBoost is the most well-known and first successful Boosting
algorithm for the problem of binary classification~\cite{Freund97}.
Given a training data set, the goal of AdaBoost is learning a classifier $H(\vx)$
based on a linear combination of weak classifiers, 
$G_m:X\rightarrow L\in\left\{+1,-1\right\}$,
to produce a powerful ``committee'' 
\begin{equation}
 h(\vx)=\sum^M_{m = 1} \beta_m G_m(\vx) \; ,
\end{equation}
whose prediction is $H(\vx) = \sign(h(\vx))$.
It can also be interpreted as a stage-wise algorithm fitting an additive 
model~\cite{Friedman00}.
This interpretation provides, at each round \textit{m}, a \emph{direction} for 
classification, $G_m(\vx) = \pm 1$, and a \emph{step size}, $\beta_m$, the 
former understood as a direction on a line and the latter as a measure of 
confidence in the predictions of $G_m$.
Both elements are estimated in such a way that they minimize the 
\emph{Binary Exponential Loss function} (BEL) \cite{Friedman00,Schapire99}
\begin{equation}
{\cL}\left(l,G_m(\vx)\right) =\exp(-l\,G_m(\vx )),
\end{equation}
defined on the value of $z =l\,G_m(\vx )$, usually known as the
\emph{margin} \cite{Allwein00,Zou08}.  To achieve this, a weight
distribution on the training set assigns to each training sample
$\vx_n$ a weight $w(n)$.  At iteration $m$, the algorithm adds to the
ensemble the best weak-learner according to the weight distribution.
The training data updates its weights taking into account
$\cL(l,\beta_m G_m(\vx))$.  Miss-classified samples increase their
weights while those of correctly classified are decreased.  In this
way, new weak learners concentrate on difficult un-learnt parts of the
data-set. Since there are two margin values $z= \pm 1$,  only two
weight updates, $\omega(n) = \omega(n) e^{\pm\beta_m}$, may
be achieved at each iteration.  In Section~\ref{Sec:Costs} we
introduce a vectorial encoding that provides a margin allowing various
weight updates depending on the cost associated to the classifier
response.


\subsection{Multi-class Boosting with multi-class responses}
\label{Subsubsec:Margin}

With the emergence of AdaBoost, extensions to multi-class problems
also appeared.  There are a large number of Boosting algorithms
dealing with this type of classification.  For ease of cataloging, we
divide them into two groups: algorithms that \emph{decompose the
  problem into binary sub-problems} (thus, using binary weak learners)
and algorithms that \emph{work simultaneously with all labels} (using
multi-class weak learners or computing a posteriori probabilities at
the same iteration).  This section describes in detail the second
approach, that we adopt for our proposal.

The AdaBoost.M1 algorithm proposed by Freund and Schapire uses
multi-class weak learners while maintaining the same structure of the
original AdaBoost~\cite{Freund96}. The main drawback of this approach
is the need for ``strong learners'', i.e.  hypotheses that can achieve
an accuracy of at least $50\%$. This requirement may be too strong
when the number of labels is high.  A second approach is the
multi-class version of LogitBoost~\cite{Friedman00}.  As its binary
counterpart, it estimates separately the probability of membership to
each label based on a multi-logit parametrization.

The contributions most directly related to our proposal are those grounded on a 
vectorial insight.
We want to generalize the symmetry of class-label representation in the 
binary case to the multi-class case. A successful way of achieving this goal
is using a set of vector-valued class codes 
that represent the correspondence between the label set $L = \{ 1,\ldots ,K \}$ 
and a collection of vectors $Y = \left\{\vy_1,\ldots,\vy_K \right\}$.  
Vector $\vy_l$ has a value 1 in the $l$-th coordinate and $\frac{-1}{K-1}$ 
elsewhere.
So, if $l=1$, the code vector representing class $1$ is $\vy_1 = \left(1, 
\frac{-1}{K-1},\ldots,\frac{-1}{K-1} \right)^\top$.
It is immediate to see the equivalence between classifiers $H(\vx)$ defined over 
$L$ and classifiers $\vf(\vx)$ defined over $Y$:
\begin{equation}
\label{eq:correspondencia_1}
 H(\vx) = l \in L  \ \Leftrightarrow \  \vf (\vx) = \vy_l \in Y \: .
\end{equation}
This codification was first introduced by Lee, Lin and
Wahba~\cite{Lee04} for extending the binary Support Vector Machine to
the multi-class case.  Later Zou, Zhu and Hastie generalized
the concept of binary margin to the multi-class case using a related
vectorial codification~\cite{Zou08}.  A $K$-dimensional vector $\vy$
is said to be a \emph{margin vector} if it satisfies the
\emph{sum-to-zero} condition $\sum^K_{k = 1}y(k) = 0$.  In other
words, $\vy^\top\vone=0$, where $\vone$ denotes a $K$-dimensional
vector of ones.  Margin vectors are useful for multi-class
classification problems for many reasons.  One of them comes directly
from the sum-to-zero property.  It is known that, in general, every
vectorial classifier
$\textbf{f}(\vx) = (f_1 (\vx), \ldots , f_K (\vx))^\top$ has a direct
translation into a posteriori probabilities
$P(l=k\mid\vx), \forall k \in L ,$ via the Multi-class Logistic
Regression Function,
\begin{equation}
P(l = k \mid \vx) = \frac{ \exp(f_k (\vx)) }{\sum^K_{i=1} \exp(f_i (\vx)) } \: .
\end{equation}
However, $\vf(\vx)$ produces the same posterior probabilities as
$\vg(\vx) = \vf(\vx) + \alpha(\vx)\cdot \textbf{1}$, where
$\alpha(\vx)$ is a real-valued function.  Such is the case, for
example, when $\alpha(\vx) = -f_K (\vx)$. If $\vf(\vx)$ is a margin
vector, then we can define
the equivalence relation
$\vf(\vx) \sim \vg(\vx) \Leftrightarrow \exists \alpha : \vX \mapsto
\Re \mid \vg(\vx) = \vf(\vx) + \alpha(\vx)\textbf{1}$ on 
the set of functions
$\cF = \left\{ \textbf{f} : \vX \mapsto \Re^K \right\}$.
Hence, margin functions are representatives of equivalence classes.

Using this codification, Zhu, Zou, Rosset and Hastie generalized
the original AdaBoost to multi-class problems under an statistical
point of view~\cite{Zhu09} .  Since this work is a key ingredient for
subsequent derivations, we describe the main elements upon which it is
grounded.  First, the binary margin in AdaBoost, $z = yf(\vx)$, is
replaced with the \emph{Multi-class Vectorial Margin}, defined as the
scalar product
\begin{equation}
\label{eq:margin}
z:=\vy^\top \vf(\vx) .
\end{equation}
The essence of the margin approach resides in maintaining
negative/positive values of the margin when a classifier has
respectively a failure/success.  That is, if $\vy,\vf(\vx) \in Y$; the
margin satisfies:
$z > 0 \Leftrightarrow \textbf{y} = \textbf{f}(\textbf{x})$, and
$z < 0 \Leftrightarrow \textbf{y} \neq \textbf{f}(\textbf{x})$.
The only two values for the margin are
\begin{equation}
\label{eq:marginSAMME}
z = \vy^\top\vf(\vx)  = \left\{
  \begin{array}{l l}
    \frac{K}{(K-1)} & \quad \mbox{if } \:\: \vf(\vx) = \vy \;, \\
    \frac{-K}{(K-1)^2} & \quad \mbox{if} \:\: \vf(\vx) \neq \vy \;.\\
  \end{array} \right.
\end{equation}
%
The \emph{Multi-class Exponential Loss function} (MEL) is
\begin{equation}
\label{eq:Multi_ExpoLossFun}
 \cL\left(\vy,\vf(\vx)\right) := \exp\left(-\frac{1}{K} \vy^\top\vf(\vx) \right) = \exp\left(-\frac{1}{K} z \right) \: .
\end{equation}
The presence of the constant $1/K$ is important but not determinant for the 
proper behavior of the loss function.
An interesting property of this function that explains the addition of 
the constant $1/K$ comes from the following equality:
\begin{equation}
\label{eq:MediaGeome}
\exp\left(-\frac{1}{K} \sum^K_{k=1} y(k) f_k(\vx) \right) = \left( \prod^K_{k=1}\ \exp\left( -y(k)f_k(\vx) \right) \right)^{1/K}  .                          
\end{equation}
Hence this multi-class loss function is a geometric mean of the binary 
exponential loss applied to each pair
$(\vy,\vf(\vx))$ (i.e. component-wise margins).
%
In spite of the critique in~\cite{Mukherjee10}, Zhu, Zou, Rosset, and
Hastie~\cite{Zhu09} proved that this loss function is
\emph{Fisher-consistent}~\cite{Zou08}, which means that the population
minimizer
\begin{equation}
\arg\min_{\vf} E_{Y|X = \vx} \left[ \cL\left(Y,\vf(\vx) \right)\right],
\end{equation}
has unique solution in the hyperplane of margin vectors and
corresponds to the multi-class Bayes optimal rule~\cite{Zhu09},
\begin{equation}
\arg\max_k f_k (\vx) = \arg\max_k P\left(Y = k |\vx\right).
\end{equation}

So, with enough samples we may recover the exact
Bayes rule by minimizing the MEL~(\ref{eq:Multi_ExpoLossFun}).
These results give formal guarantees for learning multi-class
classifiers.  We generalize this loss in our proposal (see
Section~\ref{Sec:Costs}).
Other loss functions, such as the \emph{logit} or $L_2$, share this
property and may also be used for building Boosting algorithms.
Similarly, Saberian and Vaconcelos justified that other margin vectors
could have been used for representing labels~\cite{Saberian11,Zou08}, and
therefore develop alternative algorithms.

SAMME~\cite{Zhu09} (Stage-wise Additive Modeling using a Multi-class
Exponential loss function) resorts to the
MEL~(\ref{eq:Multi_ExpoLossFun}) for evaluating classifications
encoded with margin vectors.  The expected loss is then minimized
using a stage-wise additive gradient descent approach.  The resulting
algorithm only differs from AdaBoost in weak learner weight computation 
$\alpha_m = \log \left( (1-Err_m)/Err_m \right) + \log (K-1)$ (SAMME adds
$\log(K-1)$).  It is immediate to prove that the classification rule
of SAMME, $H(\vx) = \arg\max_k \sum^M_{m = 1}\alpha_m I\left(T_m(\vx) = k \right)$,
is equivalent to assigning the maximum margin (\ref{eq:margin}),
$H(\vx) = \arg\max_k \vy_k^\top \vf(\vx)$, that is also directly
related to the perspective defined in \cite{Saberian11} and our
proposal (see Section~\ref{Sec:Costs}).  In the same way it is
straightforward to verify that AdaBoost becomes a special case when
$K = 2$, what makes SAMME the most natural generalization of AdaBoost
using multi-class weak-learners.

\section{Cost-sensitive Boosting}
\label{Sec:CostBoosting}

Classifiers that weigh certain types of errors more heavily than
others are called cost-sensitive.  They are used in asymmetric
classification situations, for example in medical diagnosis and object
detection problems.  For this type of problems it is usual to consider
a (\textit{K}$\times$\textit{K})-matrix $\vC$, where each entry
$C(i,j) \geq 0$ measures the cost of misclassifying an instance with
\textit{real} label $i$ when the \textit{prediction} is $j$~\cite{Elkan01}.
We expect of this matrix to have costs for correct assignments lower than 
any wrong classification, i.e. $C(i,i) < C(i,j)$, $\forall i \neq j$.
Hereafter, $\vM(j,-)$ and $\vM(-,j)$ will be used for referring to,
respectively, the $j$-th row and column vector of a matrix $\vM$.

For cost-sensitive problems a cost-dependent classification criterion is applied.
If $\vP(\vx) = (P(1|\vx), \ldots , P(K|\vx))^\top$ is the vector of a posteriori 
probabilities for a given $\vx \in X$, then the \emph{Minimum Cost Decision 
Rule} is~\cite{Elkan01}
\begin{equation} \label{eq:BayesRuleCost}
F(\vx) = arg \min_{j \in L } \vP(\vx)^\top \vC(-,j) \: ,
\end{equation}
that is the minimizer of the risk function
$\cR(\vP(\vx),\vC(-,j)) := \vP(\vx)^\top \vC(-,j)$ with respect to
$j \in L$.  When dealing with multi-class problems it is important to
understand how the consideration of a cost matrix influences the decision
boundaries. O'Brien's et al.~\cite{O'Brien08} discloses a concise
glossary of linear algebra operations on a cost matrix and their
effects on decision boundaries.  Let
\begin{equation}
\label{eq:frontera_decision}
  \vP(\vx)^\top \left[\vC(-,i) - \vC(-,j)\right] = 0
\end{equation}
be the decision boundary between classes $i$ and $j$, with $i \neq j$:
\begin{enumerate}
  \item Decision boundaries are not affected when $\vC$ is replaced by 
        $\alpha\vC$, for any $\alpha > 0$.
  \item Adding a constant to row $\vC(k,-)$ maintains the result unaffected.
\end{enumerate}
Taking into account the last property we will assume without loss of
generality that $C(i,i) = 0 , \forall i \in L,$ i.e. the cost of
correct classifications is zero.  We will denote $0|1$\emph{-matrix}
to that with zeros in its diagonal and ones elsewhere, i.e.,
a matrix representing a cost-insensitive multi-class problem.

In the following we consider essentially asymmetric matrices.  This
regular case is the appropriate for situations where some errors are
more important than others. In other words, if $C(i,j) > C(j,i)$,
$i\neq j$, we want to push the boundary between classes $i$ and $j$
towards $j$. 
In graph theory this case would represent a directed complete graph
with paths of different module even between pairs of nodes (labels).
Other interesting case comes when considering a symmetric cost matrix.
Since symmetrical values are equal, the actual information lies in
comparing the costs associated to different decision boundaries.
Hence this structure is recommended for problems where some class
boundaries are more important than others.  In graph theory this case
is related to an undirected complete graph with different distances
between nodes.


\subsection{Binary Cost-sensitive Boosting}
\label{Subsec:CostBina}

Let us assume that a binary classification problem has a
$(2\times 2)$-cost matrix with non negative real values.
Let us also assume zero costs for correct classifications.
For ease of notation we will use $C_1$ and $C_2$ to denote the constants 
$C(1,2)$ and $C(2,1)$, respectively.
Initial attempts to generalize AdaBoost in this way came essentially from 
heuristic changes on specific parts of the algorithm, which are essentially 
different re-weighting schemes 
\cite{SunY07,Ting00,Viola01,SunY05}.

Later Masnadi-Shirazi and Vasconcelos formally addressed this problem
and introduced the \textit{Cost-Sensitive AdaBoost} (CS-AdaBoost)~\cite{Masnadi11}.  The
core idea behind the algorithm is replacing the original exponential
loss function by the \emph{Cost-sensitive Binary Exponential Loss
  function} (CBEL):
\begin{eqnarray}
\label{eq:Vascon_Loss}
\cL_{CS-Ada}(l,F(\vx)) = I(l=1)\exp\left(-l C_1 F(\vx)\right) + I(l=-1)\exp\left(-l C_2 F(\vx)\right).
\end{eqnarray}
With a a $0|1$-cost matrix it is clear that it becomes AdaBoost.
CS-Adaboost is then derived by fitting an additive model whose objective is to 
minimize the expected loss.
Like previous approaches, the algorithm needs to have a pool of weak 
learners from which to select the optimal one in each iteration, jointly with 
the optimal step $\beta$.
For a candidate weak learner, $g(\vx)$, they compute two constants summing up 
the weighted errors associated to instances with the same label, $b$ for label 
$1$ and $d$ for label $-1$.
Then $\beta$ becomes the only real solution to
\begin{equation}
\label{eq:Beta_CS-Boost}
\begin{split}
2 C_1 b \cosh \left( \beta C_1 \right) + 2 C_2 d \cosh \left( \beta C_2 \right) = 
T_1 C_1 \mathrm{e}^{ -\beta C_1 } + T_2 C_2 \mathrm{e}^{ -\beta C_2 } \: .
\end{split}
\end{equation}
The algorithm adds the pair $(g(\vx),\beta)$ that minimizes $\cL_{CS-Ada}(l,F(\vx) + \beta 
g(\vx))$ to the model.

On the other hand, Landesa-Vazquez and Alba-Castro's work \cite{Landesa12} 
discuss the effect of an initial non-uniform weighing of instances to endow 
AdaBoost with a cost-sensitive behavior.
The resulting method, Cost-Generalized AdaBoost, takes advantage of the 
AdaBoost's original structure.
See N. Nikolaou et al.'s paper \cite{Nikolaou16} for an excellent summary 
of cost-sensitive binary Boosting algorithms.


\subsection{Multi-class Cost-sensitive Boosting}
\label{Subsec:previos_cost}

There are several works in the literature that address the cost-sensitiveness of 
a problem in a paradigm-independent framework
\cite{Elkan01,Domingos99,Zhou10,Xia09}.  We will not consider these
cases since we are interested in introducing costs in the multi-class
Boosting context. In the following we review the contributions
conceived for this purpose.

The \emph{AdaC2.M1} algorithm~\cite{Sun06} is probably the first
including costs when using multi-class weak learners.  The idea behind
it is combining the multi-class structure of
AdaBoost.M1~\cite{Freund96}, with the weighting rule of
AdaC2~\cite{SunY05}, hence its name.
As its multi-class counterpart, it fails in computing $\alpha$-values
only available for ``not so weak'' learners. Moreover, they code
the costs of label $l$ into a
single value, $C_l = \sum^K_{j = 1} C(l,j)$, hence loosing the
structure of the cost matrix.

The $L_p$\emph{-CSB} algorithm~\cite{Lozano08} was originally
conceived to solve instance-dependent cost-sensitive problems. It
resorts to relational hypothesis, $h: X \times L \rightarrow [0,1]$,
satisfying the stochastic condition, $\sum_{l\in L} h(l|\vx) = 1$, to
solve the minimization
\begin{equation}
\label{eq:FunObje_LpCSB}
\arg\min_h \frac{1}{N} \sum^N_{n=1} C(l_n, \arg\max_k h(k|\vx_n) ).
\end{equation}
Expression (\ref{eq:FunObje_LpCSB}) can be approximated by the following convexification:
\begin{equation}
\label{eq:FunObje2_LpCSB}
\arg\min_h \frac{1}{N} \sum^N_{n=1} \sum^K_{k=1} h(k|\vx_n)^p C(l_n,k) \: ,
\end{equation}
which becomes the aim of the Boosting algorithm.
They follow an extended-data approach, just like AdaBoost.MH \cite{Schapire99}, 
for stochastic hypothesis.
A drawback when applying $L_p$-CSB comes from the selection of the optimal 
value $p$ for the norm. There is no clear foundation for selecting it.

Finally, the \emph{MultiBoost} algorithm~\cite{Wang13} resorts to
margin vectors, $\vy,\vg(\vx)\in Y$ and the loss
\begin{equation}
\label{eq:FunExpChino}
\cL( l,\vf(\vx) ) := \sum^K_{k = 1} C(l,k) \exp( f_k(\vx) ) \: ,
\end{equation}
to carry out a gradient descent search.  The loss
(\ref{eq:FunExpChino}) does not generalize any previous binary problem.


\section{BAdaCost}
\label{Sec:Costs}

In this section we introduce a new multi-class cost-sensitive margin,
based on which we derive BAdaCost. We also relate it with previous
algorithms and prove that it generalizes SAMME~\cite{Zhu09} and PIBoost
\cite{Baldera14} multi-class approaches and
CS-AdaBoost~\cite{Masnadi11}  binary cost-sensitive scheme.


\subsection{Multi-class cost-sensitive margin}
\label{Subsec:MC_CS_margin}

We assume known the $K \times K$ cost matrix $\textbf{C}$. We define 
$\textbf{C}^*$ as
\begin{equation} \label{eq:C_aste}
C^*(i,j)  = \left\{
  \begin{array}{l l}
   C(i,j) & \quad \mbox{if} \:\: i \neq j\\
   -\sum_{h = 1}^K C(j,h) & \quad \mbox{if} \:\: i = j\\
  \end{array} \right.
, \:\: \forall i,j \in L.
\end{equation}
That is, we obtain $\textbf{C}^*$ from $\textbf{C}$ by replacing the
$j$-th zero in the diagonal with the sum of the elements in the $j$-th
row with negative sign.
$C^*(j,j)$ represents a ``negative cost'' associated to a correct
classification.  In other words, elements in the diagonal should be
understood as rewards for successes.

The $j$-th row in $\textbf{C}^*$, denoted as $\textbf{C}^*(j,-)$, is a margin 
vector that encodes the cost structure associated to the $j$-th label.
We define the \emph{Multi-class Cost-sensitive Margin} for sample
$(\vx,l)$ with respect to the multi-class vectorial classifier $\vg$ 
as $z_C := \textbf{C}^* (l,-) \cdot \vg(\vx)$.
It is easy to verify that if $\vg(\vx) = \vy_i \in Y$, for a certain $i \in L$, 
then $\textbf{C}^*(l,-) \cdot \vg(\vx) = \frac{K}{K-1}\textbf{C}^*(l,i)$.
Hence, multi-class cost-sensitive margins obtained from a classifier 
$\vg : \vX \rightarrow Y$ can be computed using the label-valued analogous of 
$\vg$,  $G : \vX \rightarrow L$,
\begin{equation}
\label{eq:equiv_CostMargin1}
z_C = \textbf{C}^*(l,-) \cdot \vg(\vx) = \frac{K}{K-1}\textbf{C}^*(l,G(\vx)) .
\end{equation}
So, when considering a lineal combination of discrete classifiers,
 $H = \sum^M_{m = 1} \alpha_m \vg_m$, expression 
\begin{eqnarray}
\label{eq:equiv_CostMargin2}
z_C = \sum^M_{m = 1} \alpha_m \textbf{C}^*(l,-) \cdot \vg_m(\vx) 
    = \frac{K}{K-1}\sum^M_{m = 1} \alpha_m \textbf{C}^*(l,G_m(\vx)),
\end{eqnarray}
provides a multi-class cost-sensitive margin.  

We introduce this generalized margin in the
MEL~(\ref{eq:Multi_ExpoLossFun}) to obtain the \emph{Cost-sensitive
  Multi-Class Exponential Loss} function (\emph{CMEL}),
\begin{equation} \label{eq:lossCostFun}
\mathcal{L}_C(l,\vg(\vx)) :=  \exp(z_C) = \exp \left( \textbf{C}^*(l,-) \cdot \vg(\vx) \right) .
\end{equation}
that we optimize in our algorithm.
The new margin, $z_C$, yields negative values when classifications are
correct under the cost-sensitive point of view, and positive values
for from costly (wrong) assignments.  Hence, $\cL_C$ does not need a
negative sign in the exponent.  Moreover, the range of margin values
of $z_C$ in (\ref{eq:equiv_CostMargin1}) is broader than the $z=\pm 1$
values of AdaBoost and related to the costs incurred by the classifier
decisions.

The CMEL is a generalization of the MEL~(\ref{eq:Multi_ExpoLossFun})
and CBEL~(\ref{eq:Vascon_Loss}).  Let $\vC_{0|1}$ be the cost matrix
for a cost-insensitive multi-class problem.  Since any matrix
$\lambda \vC_{0|1}$, with $\lambda > 0$, represents the same problem
\cite{O'Brien08}, then, given (\ref{eq:marginSAMME}) it is
straightforward that $\frac{1}{K(K-1)}\vC_{0|1}$ will lead to the same
values of (\ref{eq:Multi_ExpoLossFun}) when applied on the CMEL.  In
other words, the MEL is a special case of the CMEL in a
cost-insensitive problem.  On the other hand, it is also immediate to
see that for a binary classification problem the values of $\vC^*$
lead to the CBEL~(\ref{eq:Vascon_Loss}).  Hence, it is a special case
of CMEL as well.

Vectorial classifiers, $\vf(\vx) = ( f_1(\vx), \ldots , f_K(\vx) )^\top$, fitted 
using additive models collect information in coordinate $f_k(\vx)$ that can be 
understood as a \emph{degree of confidence} for classifying sample $\vx$ into 
class $k$ and, hence, they use the max rule, $\arg \max_k f_k(\vx)$, for label 
assignment~\cite{Zhu09,Baldera14,Saberian11}.
It is immediate to prove that this criterion is equivalent to assigning the 
label that maximizes the multi-class margin, $\arg \max_k \vy_k^\top \vf(\vx)$, 
that in turn is equivalent to $\arg \min_k -\vy_k^\top \vf(\vx)$.
Since $-\vy_k^\top \vf(\vx)$ is proportional to $\vC_{0|1}^*(k,-)^\top 
\vf(\vx)$, we can make the decision rule cost-sensitive selecting
the label that provides the minimum multi-class cost-sensitive margin.
Hence, the classification rule for a classifier optimizing the multi-class 
cost-sensitive margin is 
\[\
arg \min_k \vC^*(k,-) \vf(\vx).
\]


\subsection{BAdaCost: Boosting Adapted for Cost matrix}
\label{Subsec:BAdaCost}

Here we derive BAdaCost, a multi-class cost-sensitive Boosting
algorithm, as the minimizer of the empirical expected loss of
the CMEL, $\sum_{n=1}^N \cL_C (l_n,\vf (\vx_n))$,
where $\left\{ (\vx_n ,l_n) \right\}$ is our training data.

Following the statistical interpretation of Boosting~\cite{Friedman00},
the minimization is carried out by fitting a stage-wise additive
model, $\vf(\vx) = \sum_{m=1}^M \beta_m \vg_m (\vx)$.  The weak
learner selected at each iteration $m$ consists of an optimal
step of size $\beta_m$ along the direction $\vg_m$ of the largest
descent of the expected CMEL.  In \emph{Lemma}~\ref{Lemma} we show how
to compute them.

\begin{Lemma}{Optimal $(\beta_m, \vg_m(\vx))$ for CMEL}
\label{Lemma}

Let $\vC$ be a cost matrix for a multi-class problem.
Given the additive model $\vf_m (\vx) = \vf_{m-1}(\vx) + \beta_m \vg_m(\vx)$ the 
solution to
\begin{equation}
\begin{split}
  \left( \beta_m,\vg_m (\vx)\right) =  
  \arg\min_{\beta,\vg} &\sum^N_{n=1} \exp\left( \vC^*(l_n,-)\left( \vf_{m-1}(\vx_{n}) + \beta \vg(\vx_{n}) \right) \right) 
\end{split}
\end{equation}
is the same as the solution to
\begin{equation} \label{eq:expre_minim}
\begin{split}
\left( \beta_m ,\vg_m(\vx)\right) = \arg\min_{\beta,\vg} \sum^K_{j=1} S_j \exp \left( \beta C^*(j,j) \right) 
                     + \sum^K_{j=1}\sum_{k \neq j} E_{j,k} \exp \left( \beta C^*(j,k) \right) \; ,
\end{split}
\end{equation}
where $S_j=\sum_{ \left\{ n : \vg(x_n) = l_n = j \right\} } w(n)$, 
$E_{j,k}=\sum_{ \left\{ n : l_n = j , \vg(x_n) = k \right\} } w(n)$,
and the weight of the $n$-th training instance is given by:
\begin{equation}
  w(n) = \exp \left( \vC^*(l_n,-) \sum^{m-1}_{t = 1} \beta_m \vf_m (\vx_n) \right) \: .
\end{equation}
Given a known direction $\vg$, the optimal step $\beta$ can be obtained as the 
solution to 
\begin{equation}
\label{eq:beta}
\sum^K_{j=1} \sum_{k \neq j} E_{j,k} C(j,k) A(j,k)^\beta = \sum^K_{j=1} \sum^K_{h=1} S_j C(j,h) A(j,j)^{\beta} \; ,
\end{equation}
being $A(j, k) = \exp ( C^*(j,k) )$, $\forall j,k \in L$.
Finally, given a known $\beta$, the optimal descent direction $\vg$, equivalently $G$, is given by
\begin{equation}
\label{eq:g}
\begin{split}
\arg\min_G \sum^N_{n=1} w(n) A(l_n,l_n)^{\beta} I ( G(\vx_n) = l_n )
         + \sum^N_{n=1} w(n)\sum_{k \neq l_{n}} A(l_n,k)^\beta I ( G(\vx_n) = k )\; .
\end{split}
\end{equation}
\end{Lemma}

The proof of this result is in the supplementary material.  BAdaCost
pseudo-code is shown in Algorithm \ref{alg:BAdaCost}.  Just like other
Boosting algorithms we initialize weights with a uniform distribution.
At each iteration, we add a new multi-class weak learner
$\vg_m: X \rightarrow Y$ to the additive model weighted by $\beta_m$,
a measure of the confidence in the prediction of $\vg_m$.  The optimal
weak learner that minimizes (\ref{eq:g}) is a cost-sensitive
multi-class classifier trained using the data weights, $w(n)$, and a
modified cost matrix, $\vA^\beta = \exp (\beta \vC^*)$.

\begin{algorithm}                     
\caption{: BAdaCost}          
\label{alg:BAdaCost}
 \footnotesize
\begin{algorithmic}
\STATE\textbf{1-} Initialize weights $\vw$ with $w(n) = 1/N$; for $n = 1,\ldots ,N$.
\STATE\textbf{2-} Compute matrices $\vC^*$ with equation (\ref{eq:C_aste}) and $\vA$ for $\vC$.
\STATE\textbf{3-} For $m = 1$ to $M$:
\STATE\ \ \ \ \textbf{(a)} Obtain $G_m$ solving (\ref{eq:g}) for $\beta = 1$. 
\STATE\ \ \ \ \textbf{(b)} Translate $G_m$ into $\vg_m : X \rightarrow Y$.
\STATE\ \ \ \ \textbf{(c)} Compute $E_{j,k}$ and $S_j$, $\forall j,k$; as described in Lemma \ref{Lemma}.
\STATE\ \ \ \ \textbf{(d)} Compute $\beta_m$ solving equation (\ref{eq:beta}).
\STATE\ \ \ \ \textbf{(e)} $w(n) \leftarrow w(n) \exp \left( \beta_m \vC^*(l_n,-)\vg_m (\vx_n) \right)$.
\STATE\ \ \ \ \textbf{(f)} Re-normalize vector $\vw$.
\STATE\textbf{4-} Output: $H(\vx) = \arg\min_k \vC^*(k,-) \left(\sum^M_{m=1} \beta_m \vg_m (\vx)\right)$.
\end{algorithmic}
\end{algorithm}

Unlike other Boosting algorithms~\cite{Freund97, Zhu09}, here $\vg_m$
and $\beta_m$, can not be optimized independently.  We may solve this
in a similar way to e.g.~\cite{Masnadi11,Baldera14} with a local optimization.
In practice, however, we have observed that there are no significant
differences if we estimate $\vg_m$ for a fixed $\beta_m=1$
and then, given the optimal $\vg_m$,  estimate $\beta_m$.
This may be caused by the greedy nature of Boosting. Thus, we proceed
as described in Algorithm \ref{alg:BAdaCost}.


\subsection{Direct generalizations} 
\label{Subsec:generaliza_cost}

BAdaCost is a natural generalization of previous Boosting algorithms.
Here we prove that the multi-class classification algorithms
SAMME~\cite{Zhu09} and PIBoost~\cite{Baldera14} are specializations of
BAdaCost for a cost-insensitive situation. Similarly, we also prove
that CS-AdaBoost~\cite{Masnadi11} is a special case of
BAdaCost for a binary cost-sensitive problem.
The proofs are in the supplementary material.

\begin{Coro}[SAMME~\cite{Zhu09} is a special case of BAdaCost]
\label{Coro1}

When $C(i,j) = \frac{1}{K(K-1)}$, $\forall i \neq j$, then the above result is equivalent to SAMME. 
The update for the additive model $\: \vf_m(\vx) = \vf_{m-1}(\vx) + \beta_m\vg_m(\vx)$ is given by:
\[
\left( \beta_m,\vg_m(\vx )\right) = 
\arg\min_{\beta,\vg} \sum^N_{n=1} \exp \left( -\vy^\top_n \left( \vf_{m-1}( \vx_n ) + \beta \vg( \vx_n ) \right) \right) 
\]
and both optimal parameters can be computed in the following way:
\begin{itemize}
  \item[] $\vg_m = \arg\min_\vg \sum^N_{n=1} w(n) I\left(  \vg(\vx_n) \neq \vy_n \right)$
  \item[] $\beta_m = \frac{(K-1)^2}{K} \left( \log\left(\frac{1-E}{E}\right) + \log\left( K-1 \right) \right)$,
\end{itemize}
where $E$ is the sum of all weighted errors.
\end{Coro} 

\begin{Coro}[CS-AdaBoost~\cite{Masnadi11} is a special case of BAdaCost]
\label{Coro2} 

When $K = 2$ the Lemma \ref{Lemma} is equivalent to the Cost-sensitive AdaBoost.
If we denote $C(1,2) = C_1$ and $C(2,1) = C_2$, the update (\ref{eq:expre_minim}) for the additive model $F_m (\vx) = F_{m-1}(\vx) + \beta_m G_m (\vx)$ becomes:
\[
\left( \beta_m , G_m (\vx) \right) = \arg\min_{\beta,G} \sum_{ \lbrace l_n=1 \rbrace } w(n) \exp \left( -C_1 \beta G ( \vx_{n} ) \right) 
                                                        + \sum_{ \lbrace l_n=2 \rbrace } w(n) \exp \left( C_2 \beta G( \vx_n ) \right) \: .
\]
For a certain value $\beta$ the optimal direction $G_m( \vx )$ is given by
\[
\arg\min_G \left( \mathrm{e}^{ \beta C_1 } - \mathrm{e}^{ -\beta C_1 } \right) b + \mathrm{e}^{ -\beta C_1 } T_1 
           + \left( \mathrm{e}^{ \beta C_2 } - \mathrm{e}^{-\beta C_2 } \right) d + \mathrm{e}^{ -\beta C_2 } T_2 \: ,
\]
being\footnote{Here we adopt the notation used in~\cite{Masnadi11}.} : $T_1 = \sum_{ \{ n : l_n =  1 \} } w(n)$, $T_2 = \sum_{ \{ n : l_n = 2 \} } w(n)$, $b = \sum_{ \{ n : G(\vx_n) \neq l_n = 1 \} } w(n)$ and $d = \sum_{ \{ n : G(\vx_n) \neq l_n = 2 \} } w(n)$.
Given a known direction, $G(\vx)$, the optimal step $\beta_m$ can be calculated as the solution to
\[
2 C_1 b \cosh \left( \beta C_1 \right) + 2 C_2 d \cosh \left( \beta C_2 \right) = 
T_1 C_1 \mathrm{e}^{ -\beta C_1 } + T_2 C_2 \mathrm{e}^{ -\beta C_2 }\; .
\]
\end{Coro}

\begin{Coro}[PIBoost~\cite{Baldera14} is a special case of BAdaCost]
\label{Coro3}

When using margin vectors to separate a group of $s$-labels, $S \in \cP(L)$, from the rest, the result of the Lemma \ref{Lemma} is equivalent to PIBoost.
The update for each additive model built in this fashion, $\vf_m(\vx) = \vf_{m-1}(\vx) + \beta_m \vg_m(\vx)$, becomes:
\[
\left( \beta_m,\vg_m (\vx)\right) = \arg\min_{\beta,\vg} \sum_{ n=1 }^N w(n) \exp \left( \frac{-\beta}{K} \vy_n^{\top} \vg( \vx_n )  \right) \: .
\]
For a certain value $\beta$ the optimal direction $\vg_m( \vx )$ is given by 
\[
\begin{split}
\arg\min_{\vg} \left( \mathrm{e}^{ \frac{\beta}{s(K-1)} } - \mathrm{e}^{ \frac{-\beta}{s(K-1)} } \right) E_1 + \mathrm{e}^{ \frac{-\beta}{s(K-1)} } A_1 \\ +
\left( \mathrm{e}^{ \frac{\beta}{(K-s)(K-1)} } - \mathrm{e}^{ \frac{-\beta}{(K-s)(K-1)} } \right) E_2 + \mathrm{e}^{ \frac{-\beta}{(K-s)(K-1)} } A_2 \; ,
\end{split}
\]
being: 
\begin{eqnarray}
   A_1 = \sum_{ \{ n : l_n \in S \} } w(n), & & A_2 = \sum_{ \{ n : l_n \notin S \} } w(n), \nonumber \\ 
   E_1 = \sum_{ \{ n : G(x_n) \neq l_n \in S \} } \!\!\! \!\!\! \!\!\! w(n), & & E_2 = \sum_{ \{ n : G(x_n) \neq l_n \notin S \} } \!\!\! \!\!\! \!\!\! w(n) \nonumber.
\end{eqnarray}

Besides, known a direction $\vg( \vx )$, the optimal step $\beta_m$ can be calculated as
$\beta _m = s(K-s)(K-1) \log R$, where $R$ is the only real positive root of the polynomial
\[
P_m (x) = E_1 (K-s)x^{2(K-s)} + E_2 s x^K 
          - s(A_2 - E_2)x^{(K-2s)} - (K-s)(A_1 - E_1) \; .
\]
\end{Coro}


\section{Experiments} 
\label{Sec:Experiments}

In this section first we compare BAdaCost with other multi-class
cost-sensitive Boosting algorithms in the minimization of the expected
cost.  We also evaluate our approach in two asymmetric multi-class
problems such as the detection of cars and faces in arbitrary orientation.
In the experiments we use cost-sensitive tree weak-learners and
regularize our Boosting algorithm using shrinkage and feature
sampling.
 
\subsection{Minimizing costs: UCI repository}
\label{Subsec:UCI_cost}

In this test we are interested in evaluating the cost minimization
capability of the algorithm.  To this end we select 12 data sets from
the UCI repository\footnote{See Additional Material.}  that cover a
broad range of multi-class classification problems with regard to the
number of variables, labels, and data instances.

We compare BAdaCost with the algorithms presented in section
\ref{Subsec:previos_cost}: AdaC2.M1 \cite{Sun06} , $L_p$-CSB
\cite{Lozano08} and MultiBoost \cite{Wang13}.  For each data set we
proceed in the following way. First, we unify train and test data into
a single set.  Then we carry out a $5$-fold cross validation process
taking care of maintaining the original proportion of labels for each
fold.  When training, we compute a cost matrix for unbalanced problems
as done in~\cite{Baldera2015}\footnote{See Additional Material}.
Then we run $100$ iterations of each algorithm.
We resort to classification trees as base learners.
As discussed in section \ref{Subsec:previos_cost}, AdaC2.M1 and BAdaCost allow the use 
of multi-class weak learners.
MultiBoost also uses multi-class weak learners but it requires a pool 
of them to work properly.
For this reason we create a pool of $6000$ weak learners. 
We build weak-learners over sample data from $30\%$, $45\%$, and $60\%$ of the training data-set 
($2000$ weak learners for each ratio).
In third place, $L_p$-CSB translates the multi-class problem into a binary one,
thus allowing the use of binary trees.

For comparison we evaluate the average misclassification cost,
$\frac{1}{N} \sum^N_{n=1}C(l_n,H(\vx_n))$, at the end of each test.
Note that, after re-scaling the cost matrix, final costs may sum up to
a very small quantity.  We show the results in Table~\ref{Table:result_UCI}.

\begin{table}[t]
\renewcommand{\arraystretch}{0.9}
\footnotesize
\centering
{\setlength\tabcolsep{1ex} 
\begin{tabular}{l|rrrr}
                &{\bf Ada.C2M1}                    &{\bf MultiBoost}               &{\bf Lp-CSB}               &{\bf BAdaCost}\\ 
   \hline
   CarEval      &  $26\: (\pm 9)$                   &  $232\: (\pm 36)$            &  $38\: (\pm 15)$         &  $\textbf{24}\: (\pm 15)$\\
   Chess        &  $29\: (\pm 5)$                   &  $262\: (\pm 34)$            &  $\textbf{4}\: (\pm 3)$ &  $160\: (\pm 9)$\\
   Isolet       &  $289\: (\pm 48)$                 &  $140\: (\pm 15)$            &  $149\: (\pm 18)$        &  $\textbf{66}\: (\pm 14)$\\
   SatImage     &  $478\: (\pm 62)$                 &  $187\: (\pm 23)$            &  $170\: (\pm 26)$        &  $\textbf{132}\: (\pm 11)$\\
   Letter       &  $491\: (\pm 78)$                 &  $319\: (\pm 53)$            &  $161\: (\pm 23)$        &  $\textbf{66}\: (\pm 7)$\\
   Shuttle      &  $\textbf{2.1}\: (\pm 0.07)$      &  $8.9\: (\pm 0.08)$          &  $3.5\: (\pm 0.13)$      &  $3.9\: (\pm 0.3)$\\
   Cont.Meth.   &  $980\: (\pm 129)$                &  $1058\: (\pm 214)$          &  $938\: (\pm 359)$       &  $\textbf{928}\: (\pm 253)$\\
   CNAE9        &  $397\: (\pm 103)$                &  $\textbf{171}\: (\pm 57)$   &  $241\: (\pm 108)$       &  $191\: (\pm 51)$\\
   OptDigits    &  $366\: (\pm 120)$                &  $134\: (\pm 27)$            &  $170\: (\pm 34)$        &  $\textbf{30}\: (\pm 8)$\\
   PenDigits    &  $326\: (\pm 29)$                 &  $193\: (\pm 34)$           &  $162\: (\pm 72)$        &  $\textbf{18}\: (\pm 5)$\\
   Segmenta.    &  $242\: (\pm 123)$                &  $154\: (\pm 48)$            &  $94\: (\pm 18)$         &  $\textbf{50}\: (\pm 30)$\\
   Waveform     &  $905\: (\pm 113)$                &  $515\: (\pm 128)$           &  $632\: (\pm 201)$       &  $\textbf{367}\: (\pm 96)$\\
\end{tabular}}
\caption{Average costs and standard deviations (in parentheses) of Ada.C2M1, MultiBoost, $L_p$-CSB, and BAdaCost for each data set ($100$ iterations) in 
  $10^{-4}$ scale. Bold values 
  represent the best result achieved for each data base.}
\label{Table:result_UCI}
\end{table}

\begin{figure}
  \centering
  \includegraphics[width=0.6\columnwidth]{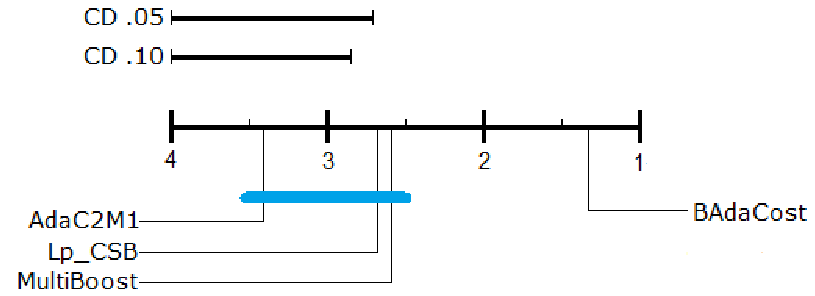}
  \caption{Comparison of ranks through the Bonferroni-Dunn test. BAdaCost's average 
           rank is taken as reference. Algorithms significantly worse than our method for 
	   a significance level of $0.10$ are unified with a blue line.}
  \label{fig:test_hipote}
\end{figure}

%
BAdaCost outperforms the rest of algorithms in most of the data sets.
To assess the statistical significance of the performance differences
among the four methods we use the Friedman test of average ranks.  The
statistic supports clearly the alternative hypothesis, i.e. algorithms
do not achieve equivalent results.  Then a post-hoc analysis
complements our arguments.  We carry out the Bonferroni-Dunn test for
both significance levels $\alpha = 0.05$ and $\alpha = 0.10$.  The
confidence distances\footnote{These values depend on the number of
  classifiers being compared jointly with the number of data sets over
  which the comparison is carried out. See \cite{Demsar06}.}  for
these tests are $CD_{0.05} = 1.2617$ and $CD_{0.10} = 1.1216$,
respectively.  Fig. \ref{fig:test_hipote} shows the final result.
%
%
We can conclude that BAdaCost is significantly better than the
AdaC2.M1 and $L_p$-CSB algorithms for the above levels of
significance. In the case of MultiBoost we can state the same
conclusion for $\alpha = 0.10$, but not for $\alpha = 0.05$ (the
difference between ranks is $1.25$).


\subsection{Multi-class object detection}
\label{Subsec:Detection}

The Boosting approach has received much attention
because it achieves state-of-the-art performance in various object
detection problems such as pedestrians~\cite{Zhang15},
multi-view faces~\cite{Mathias14} or multi-view
cars~\cite{Ohn-Bar15}. The key for the success is the use of the
feature selection capabilities of Boosting together with some robust
image descriptions such as the \emph{channel
features}~\cite{Dollar14}, pooling methods~\cite{Zhang15},
or even using ConvNets channels~\cite{Yang15}.
The usual framework uses binary classification: AdaBoost, GentleBoost,
 RealBoost, etc. In this regard, essentially multi-class detection 
 problems, such as face detection~\cite{Mathias14} or car 
detection~\cite{Ohn-Bar15}, are usually solved with a binary Boosting
 classifier per positive class (i.e. One-vs-Background strategy).  
Our goal here is to highlight the advantages of using a multi-class
cost-sensitive Boosting algorithm such as BAdaCost.

In this section we adopt a sliding window object detection approach.
There are a number of advantages in using BAdaCost for this task:
\begin{itemize}
\item \textbf{Learn the boundaries between pairs of positive classes}.
  In multi-view detection problems it is important to avoid errors
  between extreme views (e.g. a frontal and a size view car).  With
  BAdaCost we can adjust errors between views (positive classes) using
  the cost matrix. This could not be achieved under the traditional
  binary perspective.
\item \textbf{A $K$-classes classifier is faster than $K$ binary
    ones.} For example, Mathias et al.~\cite{Mathias14} used 22 face
  binary detectors in \emph{Headhunter} whereas Ohn-Bar
  et al.~\cite{Ohn-Bar15} used 75 binary detectors for car
  detection. In the experiments we will show that BAdaCost can reach
  comparable or better performance than $K$ binary detectors with much
  less computational effort.
\item \textbf{BAdaCost allows cascade calibration.} This is a
  technique used to speed-up detection algorithms by stopping the
  evaluation of the strong classifier whenever the weighted sum of
  weak-learner responses goes below a given
  threshold~\cite{Zhang07}. In Section~\ref{Subsec:Score} we
  introduce a score that can be used for calibrating BAdaCost
  cascade.
\end{itemize}

The performance of object detection algorithms depends on a number of
details such as image normalization~\cite{Benenson13}, feature
channels~\cite{Dollar14, Nam14, Yang15} and pooling
method~\cite{Zhang15} used, occlusion handling~\cite{Mathias13},
multi-scale detection policy~\cite{Dollar14}, etc. 
The goal of our experiments is not to compete with state-of-the-art
algorithms for object detection but to show that BAdaCost provides a
number of advantages that are complementary to many other types of
improvements in cascade design, feature computation, etc.

For our experiments, we have modified Piotr Dollar's Matlab
Toolbox~\footnote{\url{https://github.com/pdollar/toolbox}} with BAdaCost.
Our modified implementation will be available online~\footnote{\url{http://www.dia.fi.upm.es/~pcr/badacost.html}}.


\subsubsection{BAdaCost positive class score computation}
\label{Subsec:Score}

It is not obvious how to compute the score in the multi-class cost-sensitive Boosting.
The classification score should be positive whenever the target belongs to
the object class and negative when the assigned class is the
background.  In our case, given $M$ trained weak learners,
$\{\vg_m (\vx)\}_{m=1}^M$, and their weights, $\{\beta_m\}_{m=1}^M$,
BAdaCost produces a margin vector,
$\vf (\vx) = \sum^M_{m=1} \beta_m \vg_m (\vx)$. So, the predicted
costs incurred when classifying sample $\vx$ in one of the $K$ classes
are
$\vc = \vC^* \vf(\vx) = (c_1, \ldots, c_K)^\top$,
and assuming that the negative class has label ``1'', 
the score of sample $\vx$ is 
\begin{equation}
  \label{eq:detection_score}
  s(\vx) = (c_1 - min(c_2, \ldots, c_K)).
\end{equation}

In the following we use (\ref{eq:detection_score}) to compute the 
required score to calibrate the BAdaCost cascade~\cite{Zhang07}.


\subsubsection{Multi-view face detection}
\label{Subsec:FaceDetect}

\begin{figure}[htbp]
  \centering
  \includegraphics[width=0.45\columnwidth]{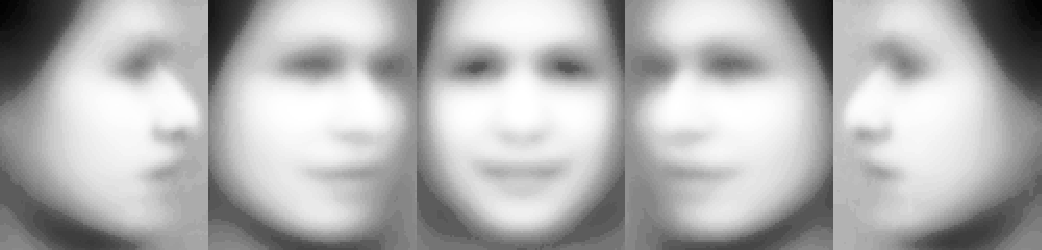}
  \caption{Mean of the AFLW training images in each face view.
           From left to right: full right profile (view 1), half right profile (view 2), 
	   frontal face (view 3), half left profile (view 4) and full left profile (view 5). }
  \label{fig:5_face_means}
\end{figure}

In this group of experiments we consider the problem of detecting faces
in images. We follow the experimental methodology of Mathias
\emph{et al.}~\cite{Mathias14}, training in the AFLW data
set~\cite{Kostinger11}, finding the classifier parameters in PASCAL
Faces, and testing in AFW~\cite{Zhu2012} and FDDB~\cite{Vidit2010}. In
our case we first run
HeadHunter\footnote{\url{https://bitbucket.org/rodrigob/doppia}}~\cite{Mathias14}
to retrieve 23073 face rectangles from AFLW.
We train our face detectors with a base size of 40 pixels.  This
procedure allows us to use the consistent face labeling of HeadHunter
to train our BAdaCost-based detector.
Second, we make $K$=5 face classes (see Fig.~\ref{fig:5_face_means})
using the AFLW data set annotations: full right profile (yaw angle less
than $-60$), half right profile (yaw angle from $-60$ to $-20$),
frontal face (yaw angle from $-20$ to $20$), half left profile (yaw
angle from $20$ to $60$) and full left profile (yaw angle greater than
$-60$). We include only images within $-35$ to $35$ degrees in
roll. Any pitch angle is also allowed.  We finally use $6136$,
$14128$, $33684$, $13483$ and $5848$ face images respectively in each
of the $5$ face orientation classes, those that are at least $40$
pixel wide. These face images include the actual image in AFLW and its
flipped version from the opposite profile view, when applicable. We
also flip the frontal view images. As negative examples we use $5772$
images without the ``Person'' label from the PASCAL VOC
2007~\cite{Everingham10} and run four rounds of hard negatives mining.

In the experiments we use the \emph{Locally Decorrelated Channel
  Features} (LDCF)~\cite{Nam14}. We have chosen to make the pyramid from one octave up
to the actual size of the input image to search for faces greater or equal to $20$ pixels.

\begin{figure}[htb]
  \centering
\subfloat[SAMME nAccNeg.]{\label{fig:PASCAL_SAMME_N_NEGS}\includegraphics[width=0.45\columnwidth] {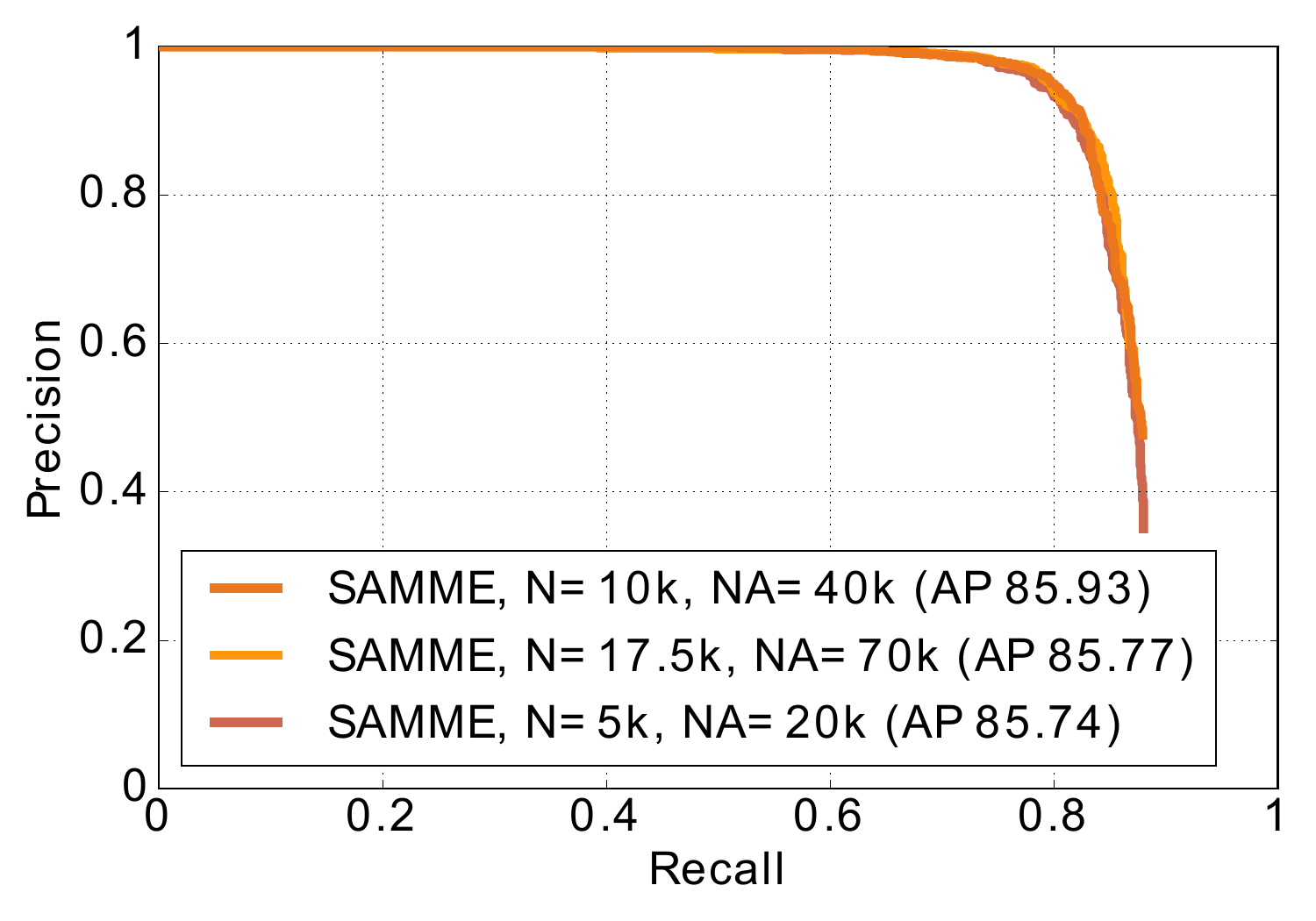}} 
\subfloat[SAMME Tree depth (D).]{\label{fig:PASCAL_SAMME_TREE_DEPTH}\includegraphics[width=0.45\columnwidth] {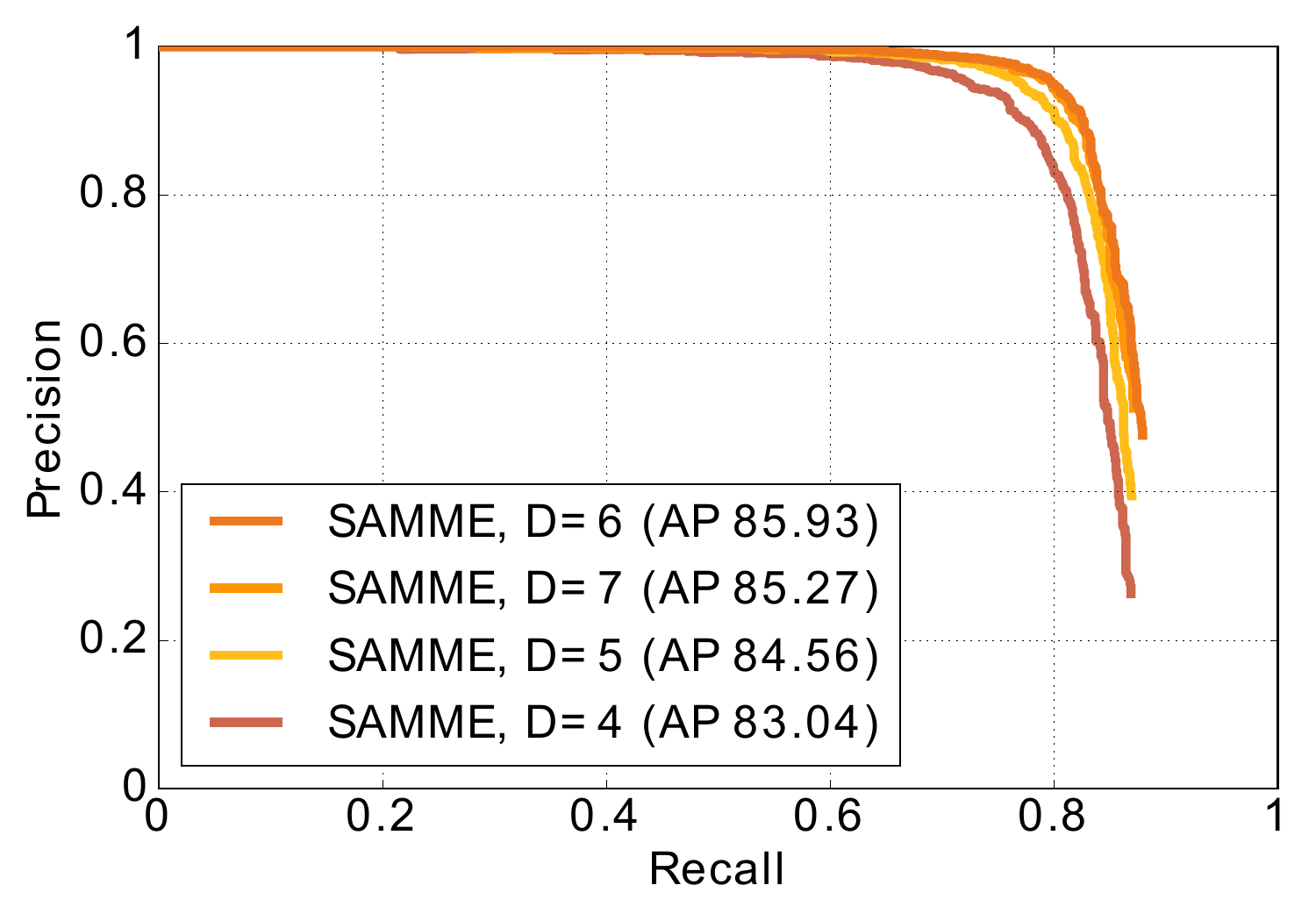}}\\
\subfloat[SAMME, SubCat, BAdaCost.]{\label{fig:PASCAL_SAMME_VS_BADACOST}\includegraphics[width=0.45\columnwidth]{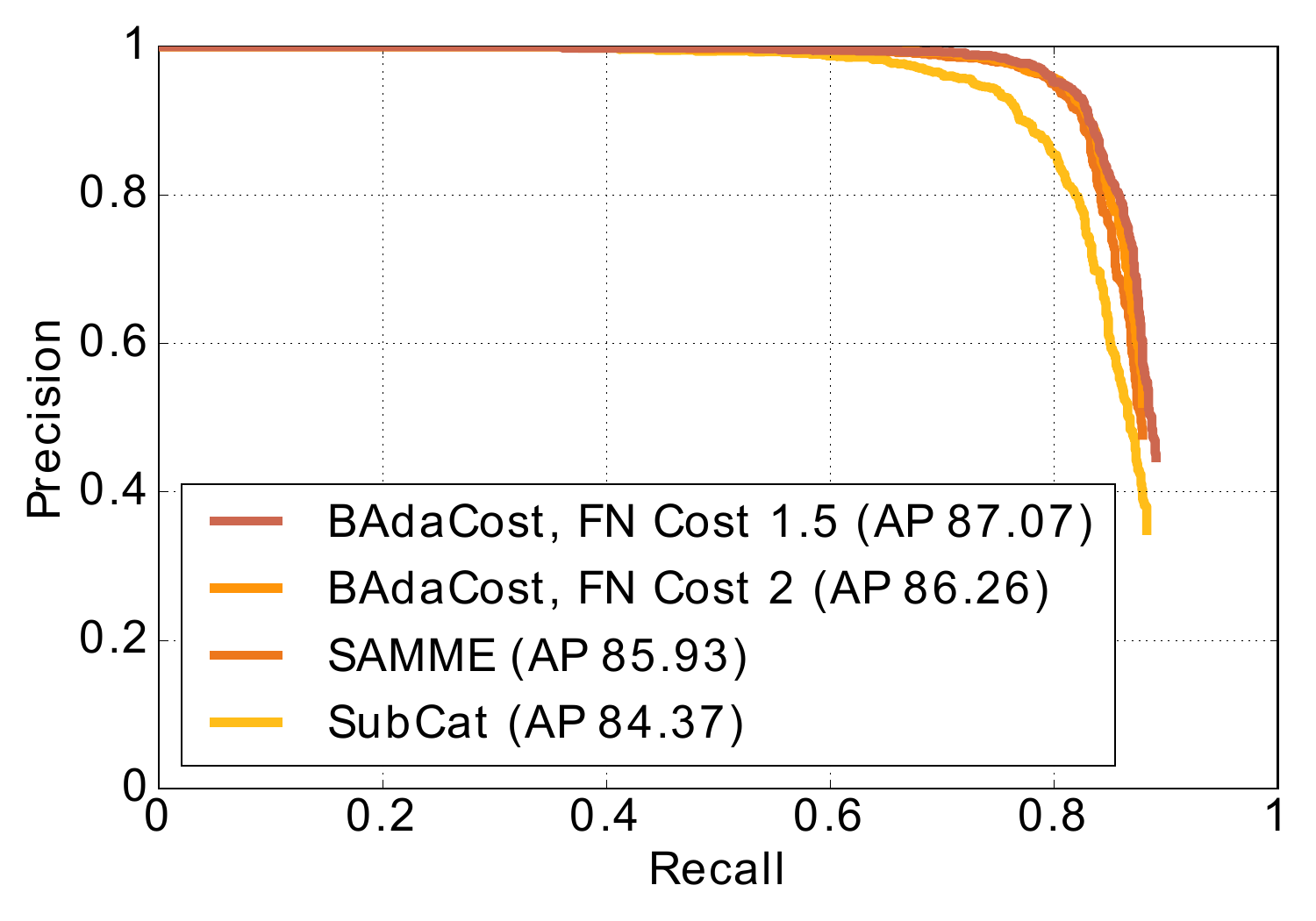}} 
\subfloat[BAdaCost \#Trees (T).]{\label{fig:PASCAL_BADACOST_ITERATIONS}\includegraphics[width=0.45\columnwidth] {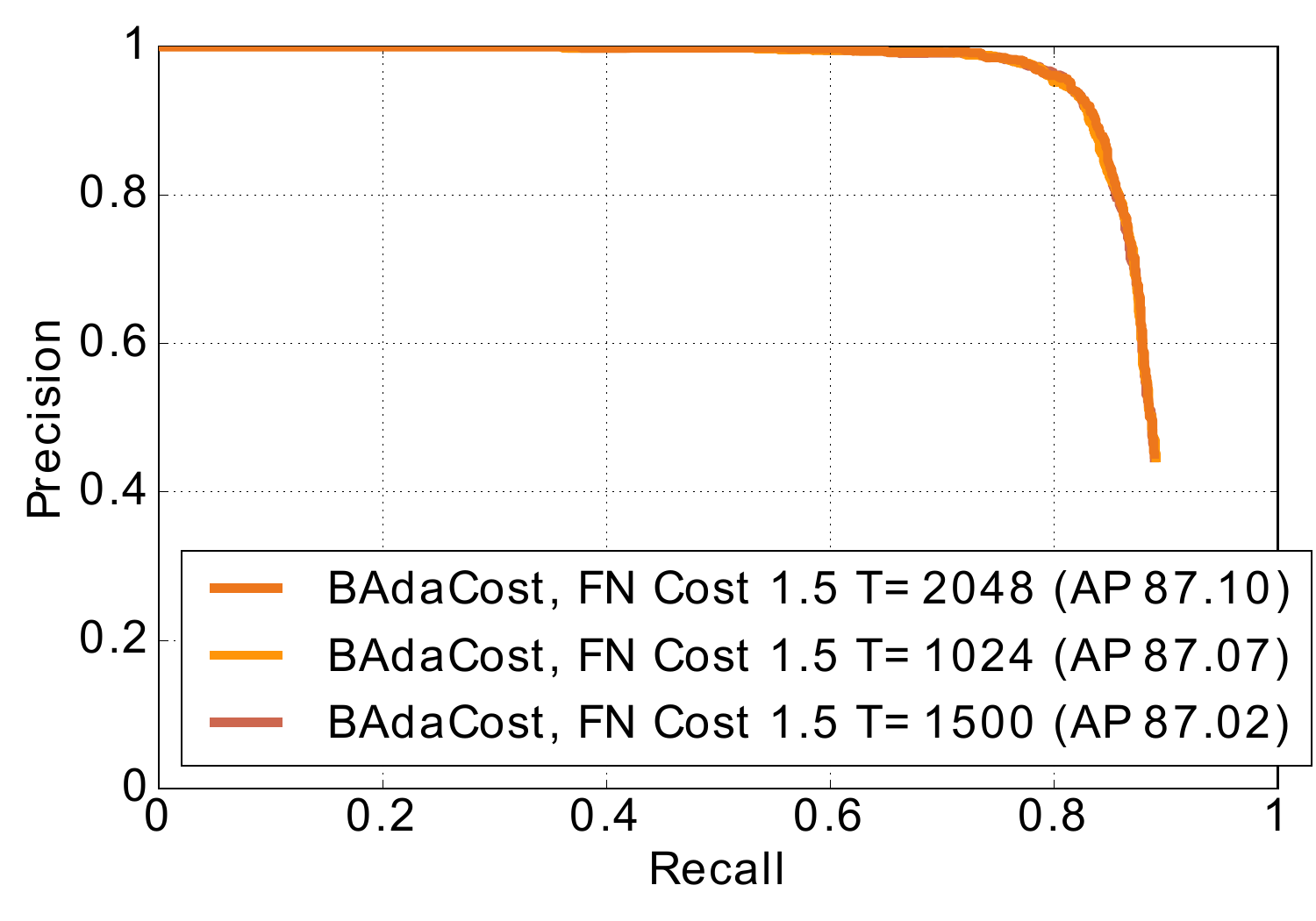}}
  \label{Search for the detector parametes training in AFLW and validating in PASCAL Faces data set.} 
  \caption{Training with AFLW and validating with PASCAL (AFLW/PASCAL).}
\end{figure}

In the following, we assign label 1 to the background class and labels 2 to 6
to the  five face orientations in Fig.~\ref{fig:5_face_means}.
To train our BAdaCost based detector we depart with a $0|1$-cost matrix (i.e SAMME). We initially set the
number of cost-sensitive trees to T=$1024$ (4 rounds with $64$, $256$,
$512$ and T weak learners, respectively) and tree depth D=$6$. We look
for the number of negatives to add (parameter $nNeg$, or $N$ for sort
in figures) per hard negative mining round and the total
amount of negatives (parameter $nAccNeg$, or \emph{NA} for sort).
In Fig.~\ref{fig:PASCAL_SAMME_N_NEGS} we show the results with
different $nAccNeg$ values training in AFLW and testing in PASCAL
Faces (AFLW/PASCAL). 
The best result in terms of Average Precision (AP) is 85.93 obtained
with $nNeg=10000$ and $nAccNeg=40000$.
After we have found the hard negative parameters, we evaluate
different tree depths (D). In Fig.~\ref{fig:PASCAL_SAMME_TREE_DEPTH}
we show the results for various Ds in the AFLW/PASCAL experiment
configuration. The best tree depth in this case is $D=6$ with AP 
85.93.

\begin{figure}[htbp]
  \centering
  \includegraphics[height=0.16\textheight]{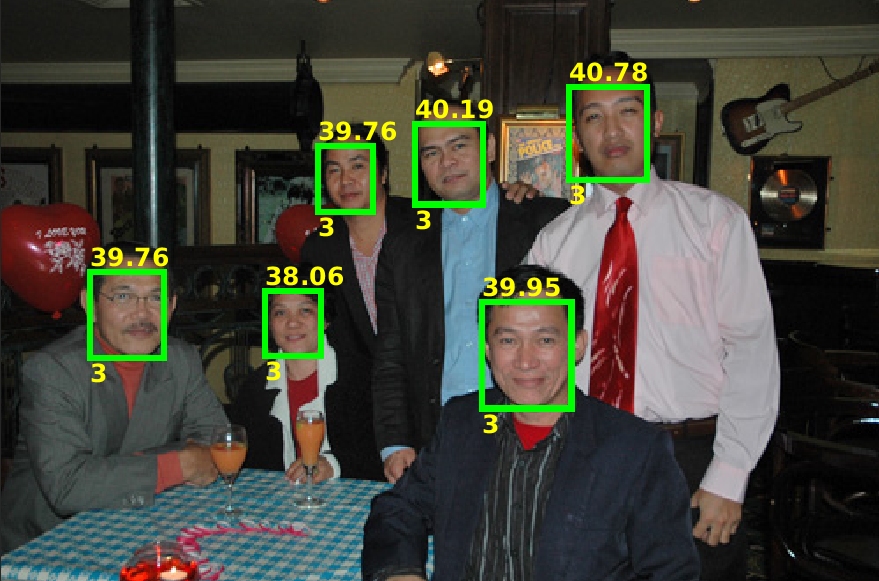} 
  \vspace{0.4ex}
  \includegraphics[height=0.16\textheight]{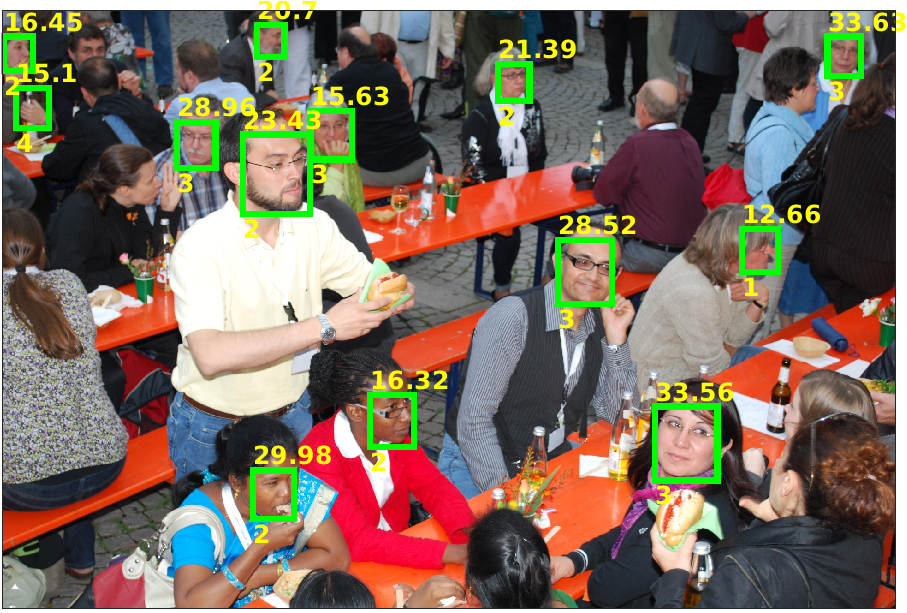} 
  \vspace{0.4ex}
  \includegraphics[height=0.16\textheight]{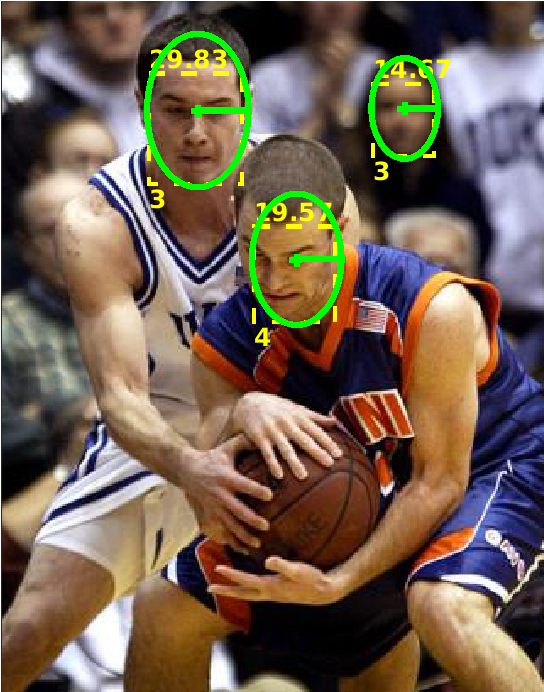}  
  \caption{Qualitative detection results for the AFLW/PASCAL
    experiment (first image), AFLW/AFW experiment (second image),
    AFLW/FDDB experiment (last image). We display faces with
    score$\geq$10.  Over each detection and under the face we show in
    yellow, respectively, the score and the estimated face class. In
    the FDDB results, we show in green the estimated ellipse from the
    yellow rectangle detection.  }
  \label{fig:FACES_QUALITATIVE}
\end{figure}

Once selected the parameters for the $0|1$-cost matrix, we can vary
the costs to adjust the boundaries between the positive classes and
background~\footnote{Note that errors between face classes in
detection do not change AP since all detections
have the same window size.}. In order to do so we define the
following cost matrix:
\begin{equation}
  \label{eq:face_costs}
  C_{\beta}=
  \left(
  \begin{array}{cc}
    0                           & \m1_{1\times 5} \\
    \beta \cdot \m1_{5\times 1} & \vC_{0|1,5\times 5}\\
  \end{array} 
  \right),
\end{equation}
where $\m1_{m\times n}$ is a ($m\times n$) matrix full of ones and 
$\vC_{0|1,n\times m}$ is the ($m\times n$) $0|1$-cost matrix.
$C_{\beta}$ assigns cost 1 to all errors between positives classes 
and $\beta$ to False Negatives (FN).
In Fig.~\ref{fig:PASCAL_SAMME_VS_BADACOST} we show the results 
obtained in the
AFLW/PASCAL experiment for values 
of $\beta=1, 1.5, 2$. We choose $\beta=1.5$ as a good compromise
giving  $50\%$ higher costs to FNs, i.e. moving the
boundaries between all positive classes towards the background
class. Finally, we test the effect of the number of trees (T) when
using $\beta=1.5$, $D=6$, $nNeg=10000$ and $nAccNeg=40000$ (see
Fig.~\ref{fig:PASCAL_BADACOST_ITERATIONS}). We find that $T=1024$
gives the optimal number of trees and AP, 87.07 in the AFLW/PASCAL
experiments (see qualitative result in first image of
Fig.~\ref{fig:FACES_QUALITATIVE}).

We compare our face detection algorithm with \emph{SquaresChnFtrs-5}
and \emph{HeadHunter}~\cite{Mathias14}. Both approaches are based on
Aggregated Channel Features
(ACF)~\cite{Dollar14}. \emph{SquaresChnFtrs-5} uses 5 face classes as
we do, but the pooling is given by summing the ACF in squares of
different sizes around the feature location.  \emph{HeadHunter} uses
21 face classes that accounts also for in plane face rotations.
Since in our experiments we use LDCF features, to have comparable
results we also train a \emph{One-vs-Background} detector per class with
LDCF. Again, we use the same face labels and images from
AFLW. We term this approach \emph{SubCat} as in~\cite{Ohn-Bar15}. We
first search for the best parameters for this approach: $nNegs=5000$
and $nAccNeg=20000$, $D=2$ and $T=4096$. 
Then we compare with the SAMME and BAdaCost approaches in
Fig.~\ref{fig:PASCAL_SAMME_VS_BADACOST}.

\begin{figure}[htbp]
  \centering
  \includegraphics[width=0.45\columnwidth]{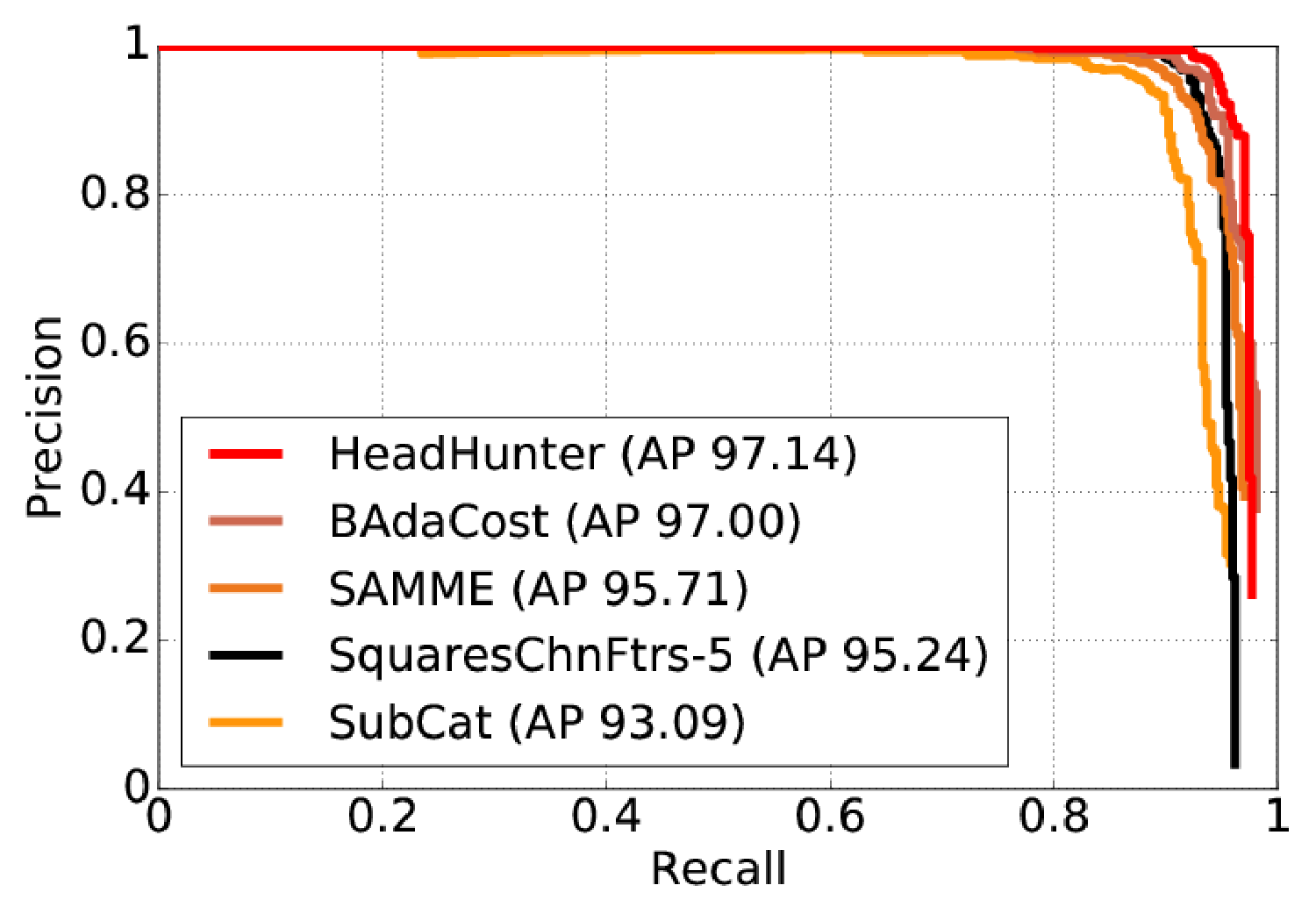}
  \caption{Training with AFLW and testing with AFW (AFLW/AFW experiment).}
  \label{fig:AFW}
\end{figure}

Following the experimentation in~\cite{Mathias14} we now test with AFW
and FDDB the best BAdaCost classifier obtained in the validation
experiments. For AFW, since all faces are wider than 40
pixels we do not scale up the input image. 
As shown in Fig.~\ref{fig:AFW} by using sensible costs (see 
(\ref{eq:face_costs})) BAdaCost can improve (i.e. learn the boundaries
between background and positive classes) the results of SAMME.
Also, the AP of BAdaCost is 97.00, very close to \emph{HeadHunter},
97.14, and better than \emph{SquaresChnFtrs-5}, 95.24. On the other
hand, \emph{SubCat}, the approach using the same features as BAdaCost,
only achieves  93.09 AP. 
See qualitative results of this experiment in the second image of
Fig.~\ref{fig:FACES_QUALITATIVE}.

\begin{figure}[htbp]
  \centering
  \subfloat[AFLW/FDDB Discrete Score results.]{\label{fig:FDDB_Disc}\includegraphics[width=0.45\columnwidth]{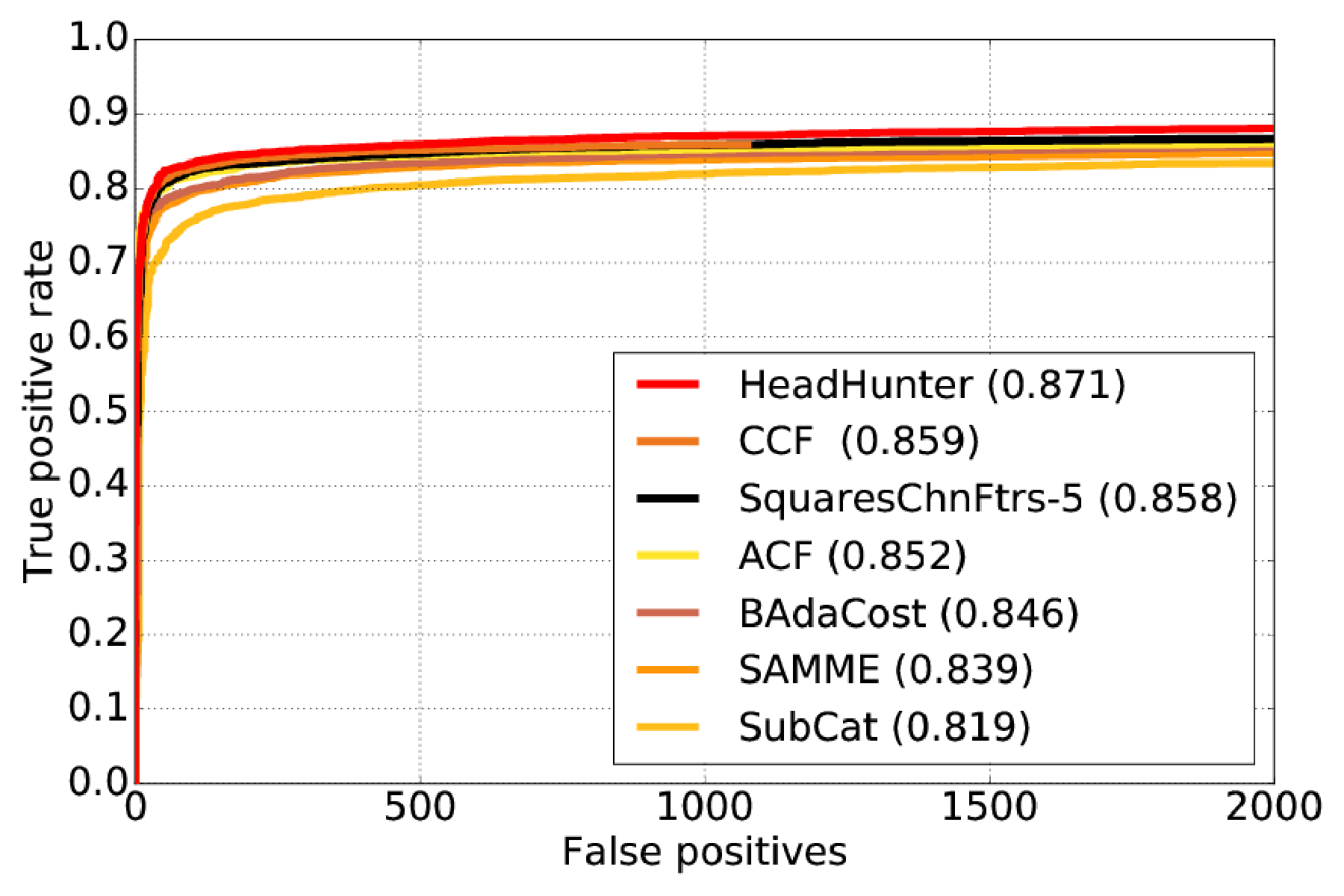}} 
  \subfloat[AFLW/FDDB Continuous Score results.]{\label{fig:FDDB_Cont}\includegraphics[width=0.45\columnwidth]{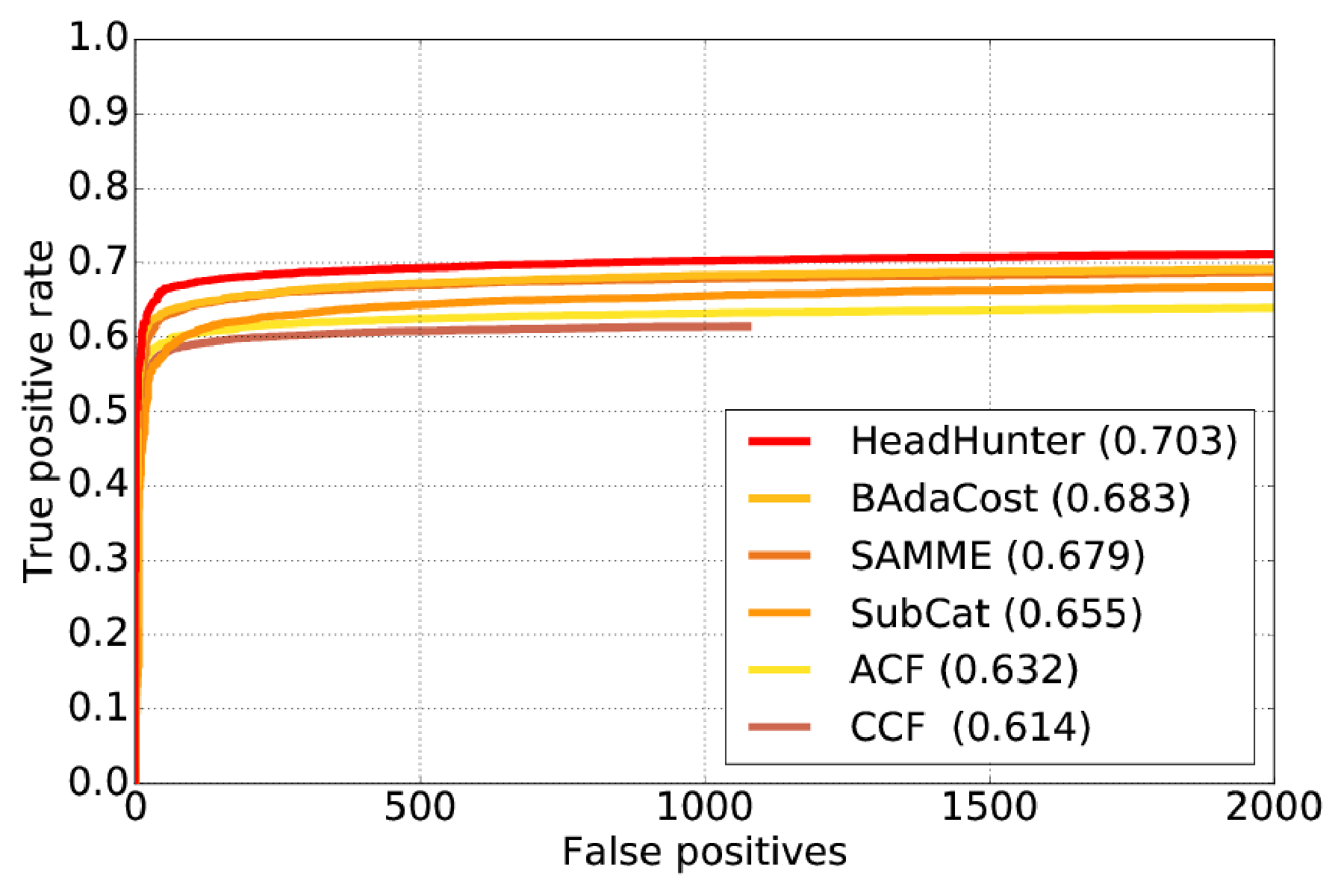}} 
  \caption{Training with AFLW and testing with FDDB.}
\end{figure}

In the FDDB experiments (results shown in Figs.~\ref{fig:FDDB_Disc}
and \ref{fig:FDDB_Cont}) we have transformed face detections into
ellipses~\cite{Mathias14}. This change enables us to achieve a better
overlap with the FDDB annotation. We show our FDDB output in the last
image of Fig.~\ref{fig:FACES_QUALITATIVE}.  In the testing protocol of
this data set there are two ROC curves: the discrete one (windows with
IoU greater than 0.5 add 1 in the True Positive Rate, TPR) and
the continuous one (windows add its IoU value in the
TPR). Unfortunately, in the latter there are no published results for
\emph{SquaresChnFtrs-5}. So we also compare ourselves with Yang et
\emph{al.}'s~\cite{Yang14} ACF based detector (\emph{ACF} label)) that
uses 6 binary detectors tuned to the yaw angle view and Convolutional
Channel Features (CCF)~\cite{Yang15}. Both approaches achieve
equivalent performance to \emph{SquaresChnFtrs-5} in the discrete score,
but have also published results for the continuous one.
In the discrete ROC experiment we again get better results than
\emph{SubCat} and \emph{SAMME}, comparable results to
\emph{SquaresChnFtrs-5}, \emph{ACF} and \emph{CCF}, and worse than
\emph{Headhunter}. In the continuous setting, \emph{Headhunter} is
the best detector,  marginally better than BAdaCost.

Our results for face detection using an out-of-the-box rigid approach like
BAdaCost are within the state-of-the-art.
\emph{HeadHunter} is the best approach with ours marginally behind
both in AFW and continuous FDDB. This is possibly due to feature
differences, the global color equalization or the large number of
face templates tuned to different facial orientations in
\emph{HeadHunter}. However, as we discuss in
section~\ref{Subsec:Efficiency}, BAdaCost is computationally much more
efficient.

On the other hand, we also get a performance improvement when Boosting multi-class
weak-learners, i.e. \emph{SAMME}, instead of binary ones,
i.e. \emph{SubCat}.  This could also be expected given that features
trained to discriminate among multiple classes are more general than
binary ones~\cite{Torralba04}.
Moreover, since in face detection all positive classes have roughly
the same aspect ratio, errors between them should not affect recall
and, hence, costs would seem useless. However, the usual way to fine
tune the negative class boundary in Boosting is by performing hard
negative mining, i.e. increasing the negative class \emph{prior}.
This can be very time consuming, especially when the classifier has a
low negative recall. Our an alternative approach is to use
costs. The improvement in performance achieved by BAdaCost vs SAMME is
due to this cost-based modification in the FP to FN ratio. Therefore,
BAdaCost allows a practical way of learning the negative class
boundary at a fixed number of hard negatives to mine.

\subsubsection{Multi-view car detection}
\label{Subsec:CarDetect}

In this section we consider the problem of detecting cars in images taken
``in the wild.'' Since car detections change their aspect ratio depending
on the viewing angle, costs play here a key role to penalize error
between positive classes. 
\begin{figure}[htbp]
  \centering
  \includegraphics[width=\columnwidth]{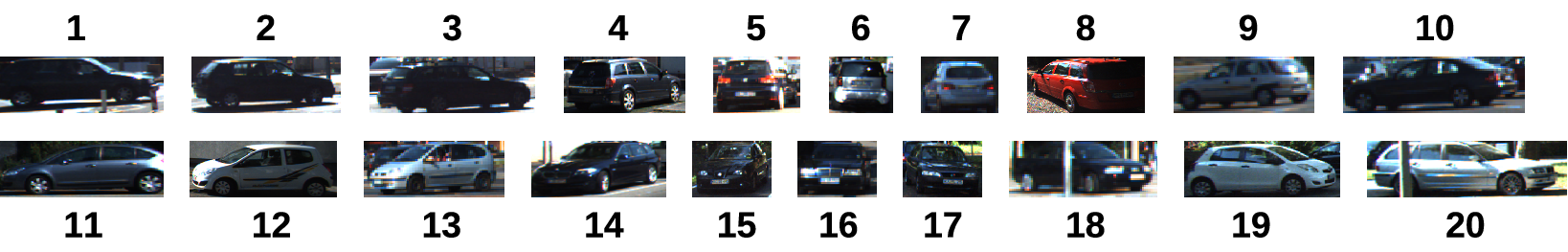}
  \caption{KITTI cars classes we use in our experiments.}
  \label{fig:KITTI_Cars_classes}
\end{figure}

We use the object detection benchmark in the KITTI
data set~\cite{Geiger2012}. 
It has three levels of difficulty: easy, moderate and hard (easy
$\subset$ moderate $\subset$ hard).  We carry out the evaluation in
each level separately and use the results in the moderate one to rank the
algorithms. In total there are 7481 images for training and 7518 for
testing. Since the testing images have no ground truth, we split the
train set in training and validation subsets: cars in the first 6733
images (90\%) to train (KITTI-train90) and the last 748 images (10\%)
as validation (KITTI-train10).

Following Ohn-Bar \emph{et al.}'s \emph{SubCat}~\cite{Ohn-Bar15},
we divide the images into $K=20$ view classes (see
Fig.~\ref{fig:KITTI_Cars_classes}) depending on the viewing angle.
\emph{SubCat} uses AdaBoost with depth-$2$ decision trees as weak
learners. For BAdaCost we use
cost-sensitive decision tree weak learners and LDCF
features~\cite{Nam14} on the standard ACF channels.  In all the
experiments with BAdaCost we train a car model of size $48 \times 84$ pixels
with a one octave-up pyramid to detect cars 24 pixel high.
Ohn-Bar et al.'s approach learns a standard Dollar AdaBoost binary
detector~\cite{Dollar14} for each car view (ignoring all other views
to extract negatives) and for three heights (25, 32 and 48 pixels).
This approach has the advantage of estimating the correct bounding box
aspect ratio.  The main disadvantages are: 1) it sweeps $20$ detectors
(at least in one scale) on each image and 2) it cannot address errors
between positive (car views) classes.  With our multi-class detector
we only go through the image once. Moreover, we can use costs to model
borders between positive classes.  Although in our approach we have to
select a fixed window shape for learning and detecting, since the
multi-class sliding window detector outcome is the view class, we can
correct the fixed size window to the mean training aspect ratio of the
predicted view class.

\begin{figure}[htbp]
  \centering
  \subfloat[SAMME nAccNeg.]
  {\label{fig:KITTI_SAMME_N_NEGS}\includegraphics[width=0.3\columnwidth]{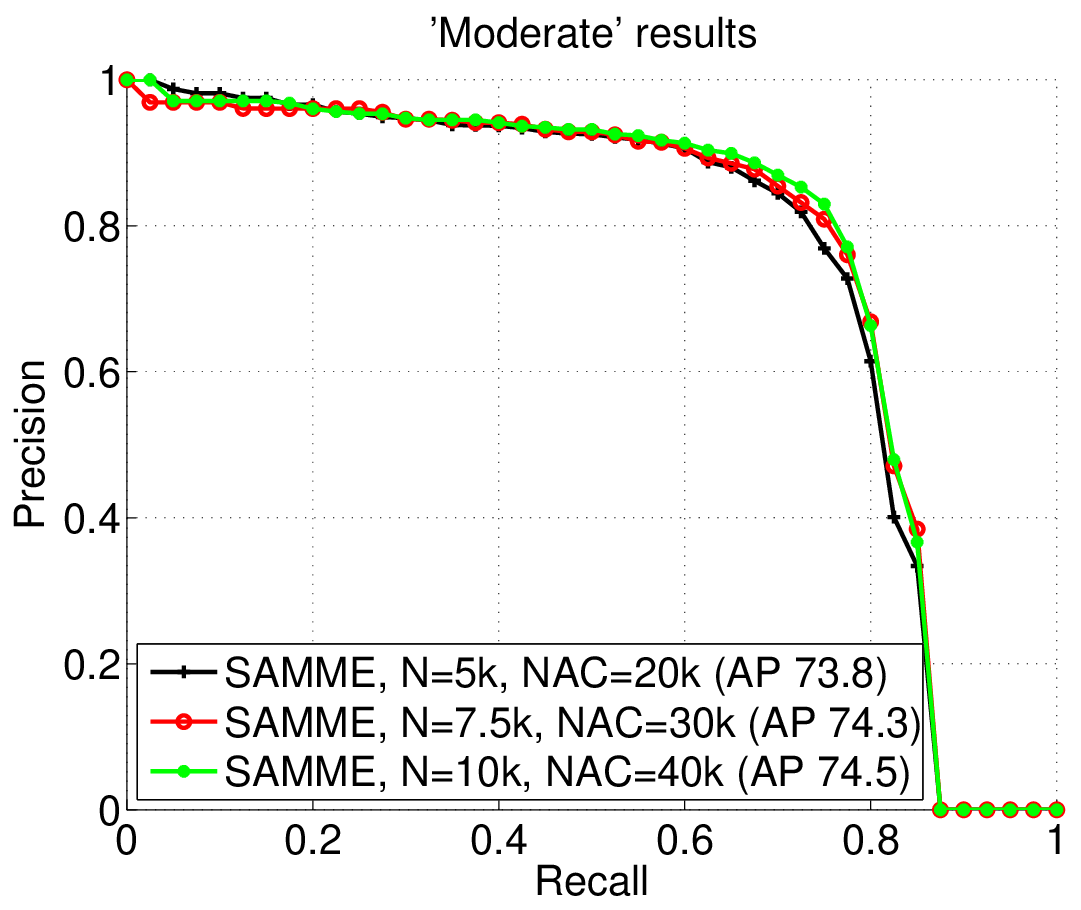}}
  \subfloat[SAMME Tree depth (D).]
  {\label{fig:KITTI_SAMME_TREE_DEPTH}\includegraphics[width=0.3\columnwidth]{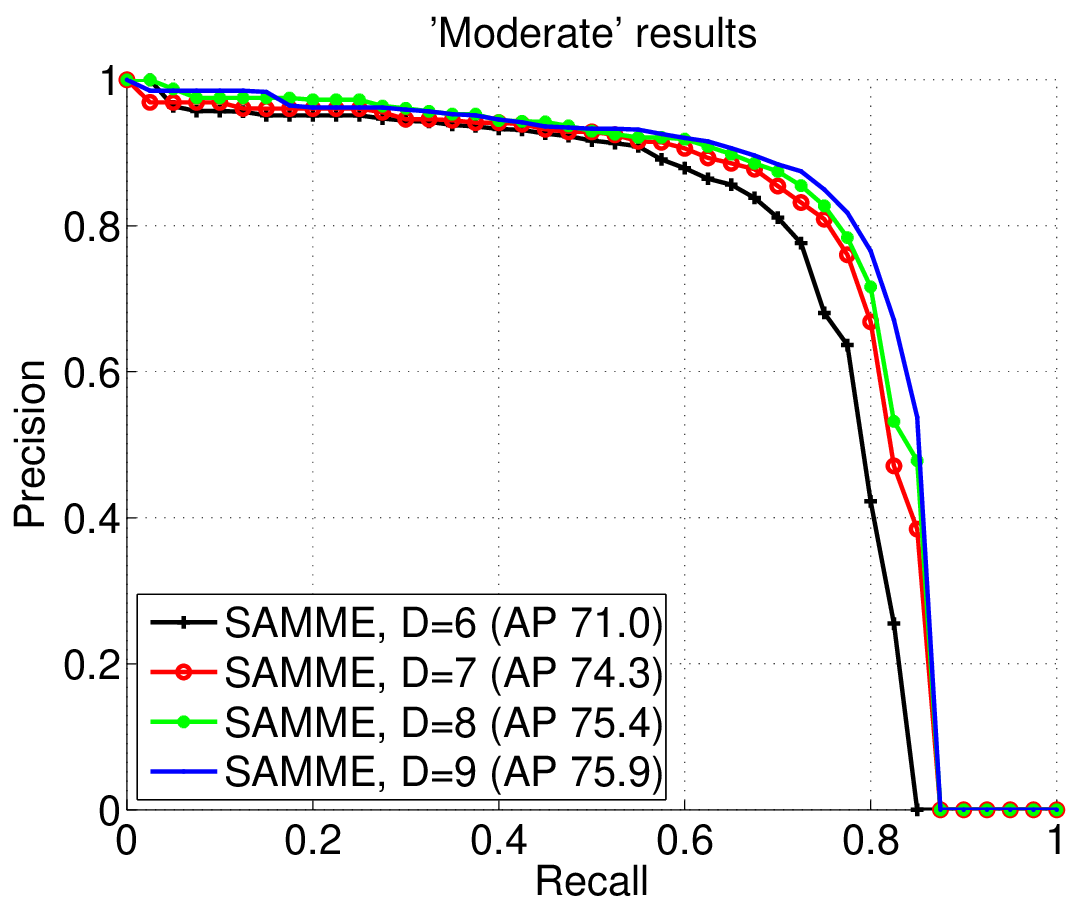}}
  \subfloat[SAMME vs BAdaCost.]
  {\label{fig:KITTI_SAMME_VS_BADACOST}\includegraphics[width=0.3\columnwidth]{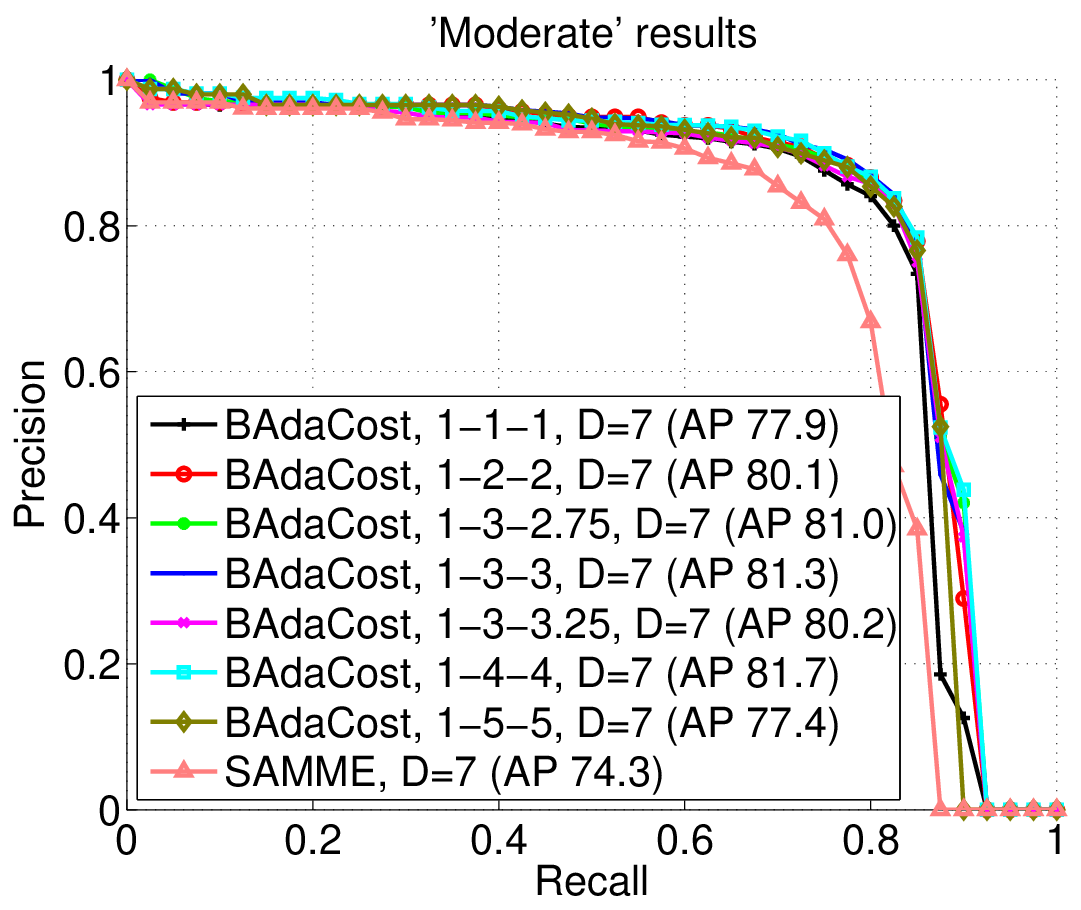}} \\ 
  \subfloat[BAdaCost T. depth (D).]
  {\label{fig:KITTI_BADACOST_TREE_DEPTH}\includegraphics[width=0.3\columnwidth]{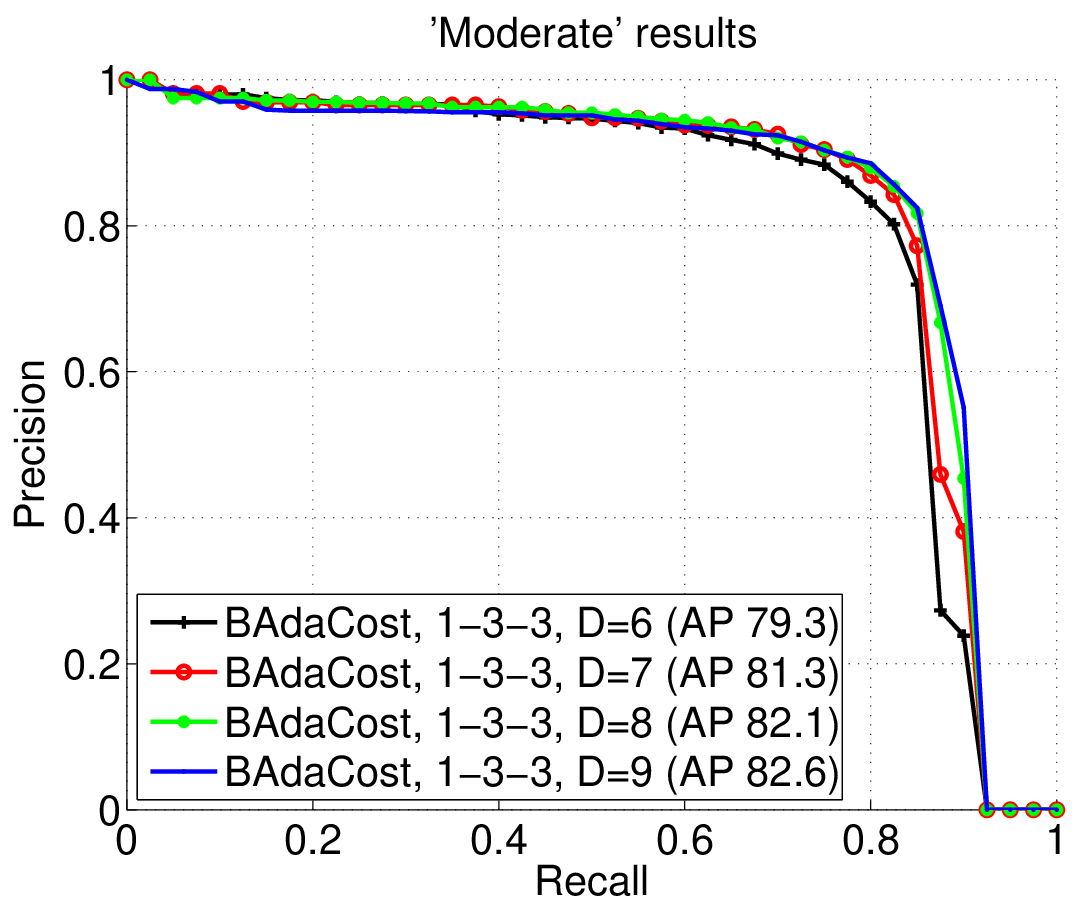}}
  \subfloat[SubCat Tree depth (D).]
  {\label{fig:KITTI_SUBCAT}\includegraphics[width=0.3\columnwidth]{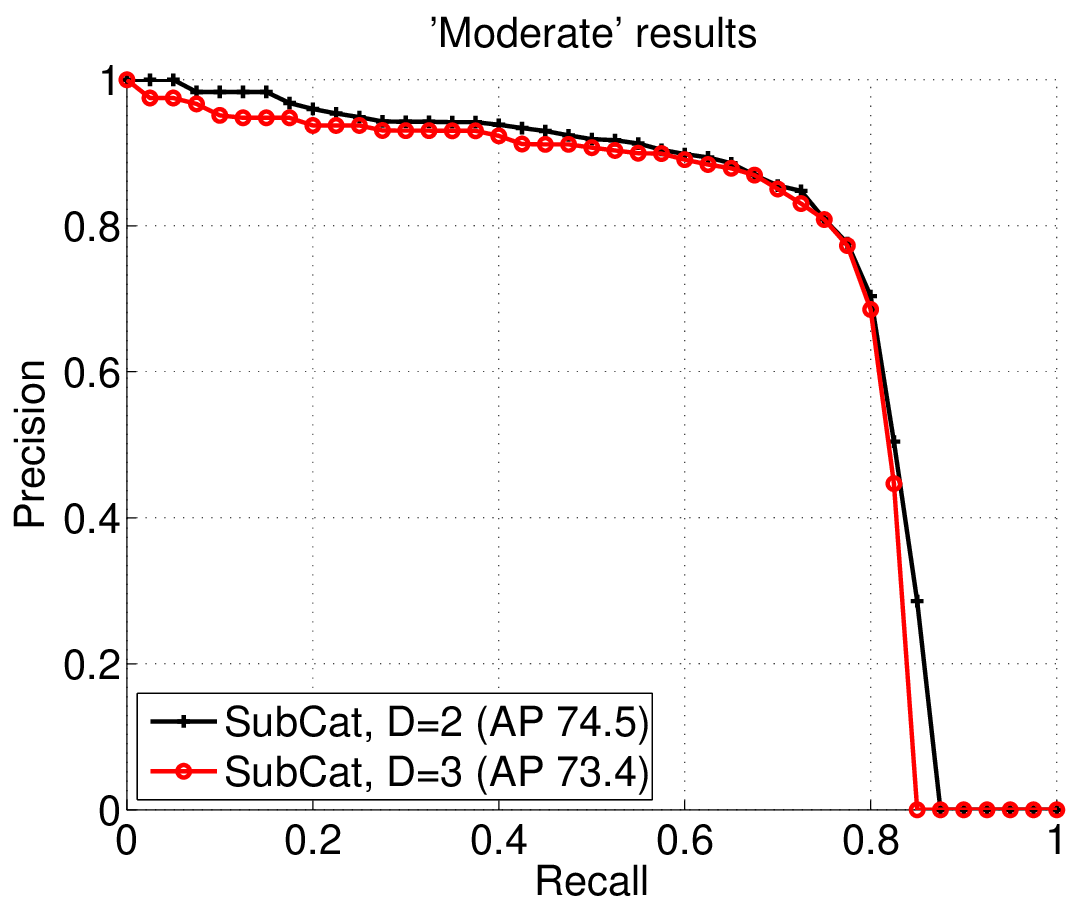}}
  \subfloat[Final comparison results.]
  {\label{fig:KITTI_SAMME_VS_BADACOST_VS_SUBCAT}\includegraphics[width=0.3\columnwidth]{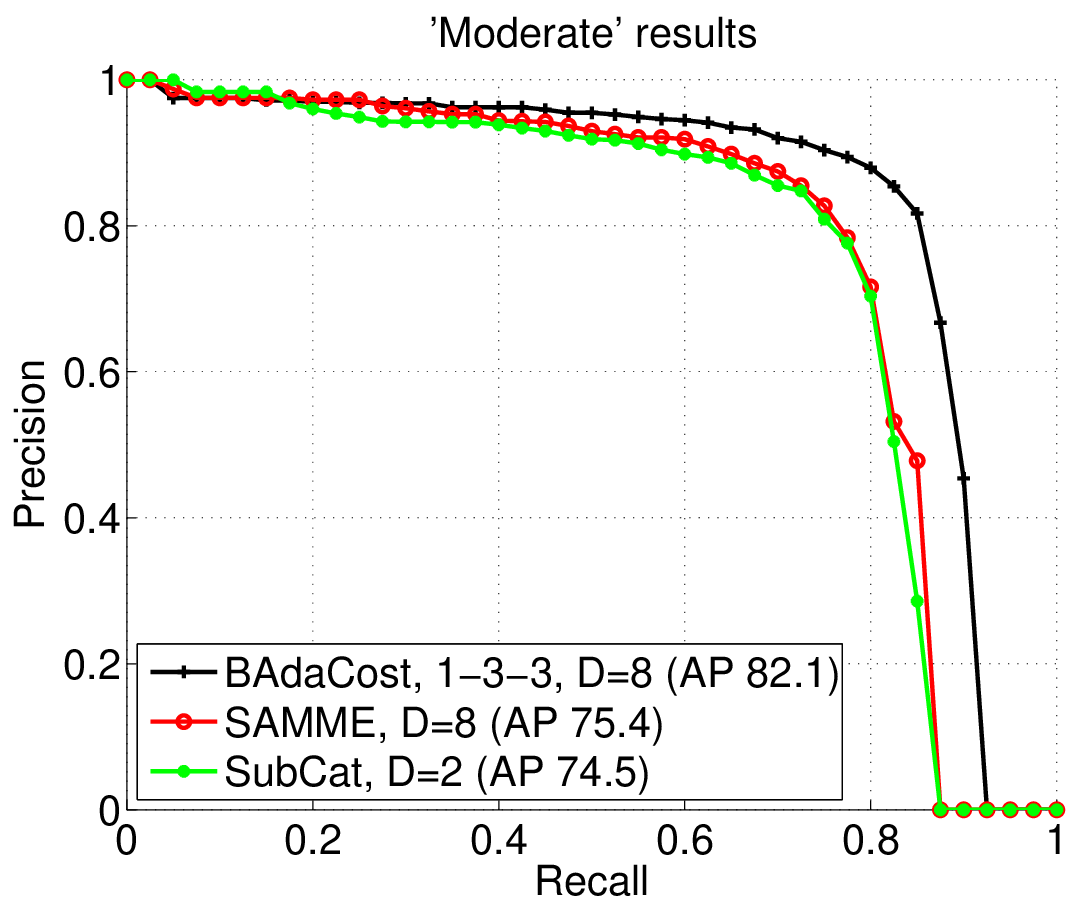}}
  \caption{Training with KITTI-train90 and validating with KITTI-train10.} 
\end{figure}

In all experiments with BAdaCost in this section we perform 4 rounds
of hard negatives mining with the KITTI training image subset
(KITTI-train90). We set the number of cost-sensitive trees to T=$1024$
(4 rounds with $32$, $128$, $256$ and T weak learners, respectively),
tree depth to D=$7$, and we look for the number of negatives per round
to add (parameter $N$ for sort) and the total amount of negatives
(parameter NA). In Fig.~\ref{fig:KITTI_SAMME_N_NEGS} we show the
results with different $nAccNeg$ values in the KITTI-train90/KITTI-10
experiments for the $0|1$-cost matrix (i.e SAMME costs). We choose
$nNeg=7500$ and $nAccNeg=30000$ since increasing it does not improve
AP. Then we evaluate different tree depths (D). In
Fig.\ref{fig:KITTI_SAMME_TREE_DEPTH} we can see that D=9, AP=75.9, is
almost equal to D=8, AP=75.4. We prefer depth $D=8$ to prevent
over-fitting.

Let $K_p$ be the number of positive classes and let us assume that
view class labels are arranged in a circular fashion like in
Fig.~\ref{fig:KITTI_Cars_classes}. We divide the cost matrix into 3
groups: FP errors (background predicted as any positive class),
FN errors (positive class classified as background) and
errors in the predicted positive class. We give weights to each of
the 3 cost groups: $\alpha$ to the FP error, $\beta$ to the FN errors
and finally $\gamma$ to the errors between positive classes.  Thus,
our cost matrix for the multi-view car detection problem is
\begin{equation}
  C_{\alpha, \beta, \gamma}=
  \left(
  \begin{array}{cc}
    0                             & \alpha \cdot \m1_{1\times 20} \\
    \beta \cdot \m1_{20\times 1} & \gamma \cdot \mP \\
  \end{array} 
  \right),
\end{equation}
where $\mP$ is a $(K_p\times K_p)$ cost matrix (i.e. main diagonal with
zeros) that assigns cost $\mP_{i,j}$, when predicting a car of class
$i$ as being of class $j$, as follows

\begin{equation}
  \label{eq:kitti_costs}
  \mP(i,j)=1-\left| \frac{2|i-j|-K_p}{K_p} \right|.
\end{equation}

Equation~(\ref{eq:kitti_costs}) assigns the lowest cost, $2/K_p$, to
close view  errors ($|i-j|=K_p-1$ or $1$) and $1$ to opposite view 
errors ($|i-j|=K_p/2$, e.g. rear to frontal class errors). Weights
$\alpha$, $\beta$ and $\gamma$ balance the relative importance between
error groups. We have tested with various combinations
of these weights. In Fig.~\ref{fig:KITTI_SAMME_VS_BADACOST} we show
different experiments labeled with $\alpha$-$\beta$-$\gamma$
values. In all of them we use the optimal value of $T=1024$,
$nNeg=7500$ and $nAccNeg=30000$. 
From the results in Fig.~\ref{fig:KITTI_SAMME_VS_BADACOST} and
Fig.~\ref{fig:KITTI_SUBCAT} we can conclude that the cost-less
multi-class approach (i.e. SAMME) and the multi-class approach with
binary detectors (i.e. One-vs-Background, SubCat) are equivalent in
terms of AP. On the other hand, when including a cost matrix to reduce
errors in the predicted view class we get noticeable improvements in
AP (e.g. from SAMME to BAdaCost 1-1-1). Moreover, since the background
has higher prior than any positive class, we need to find the right
balance between FN and $\mP$ costs in (\ref{eq:kitti_costs})
w.r.t. the FP. From the BAdaCost experiments in
Fig.~\ref{fig:KITTI_SAMME_VS_BADACOST}, we can see that by increasing
$\beta$ and $\gamma$ to 3, BAdaCost 1-3-3, we get the best results.

\begin{figure}[htb]
  \centering
  \includegraphics[width=0.49\columnwidth]{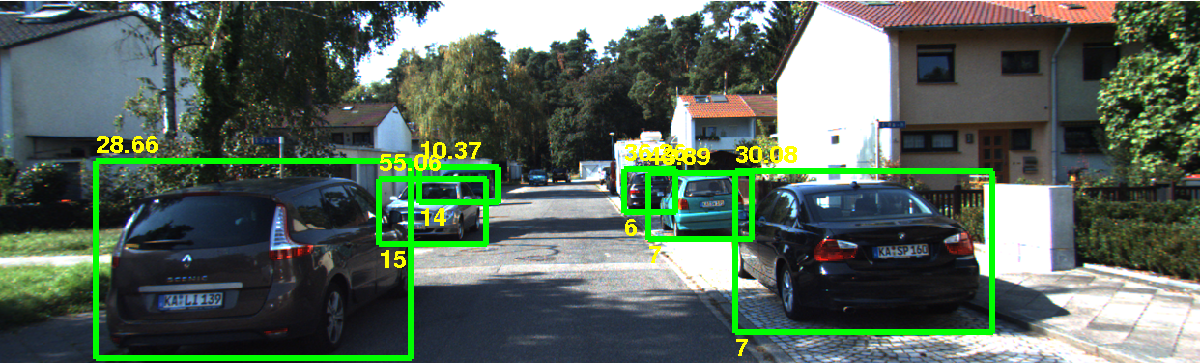}
  \includegraphics[width=0.49\columnwidth]{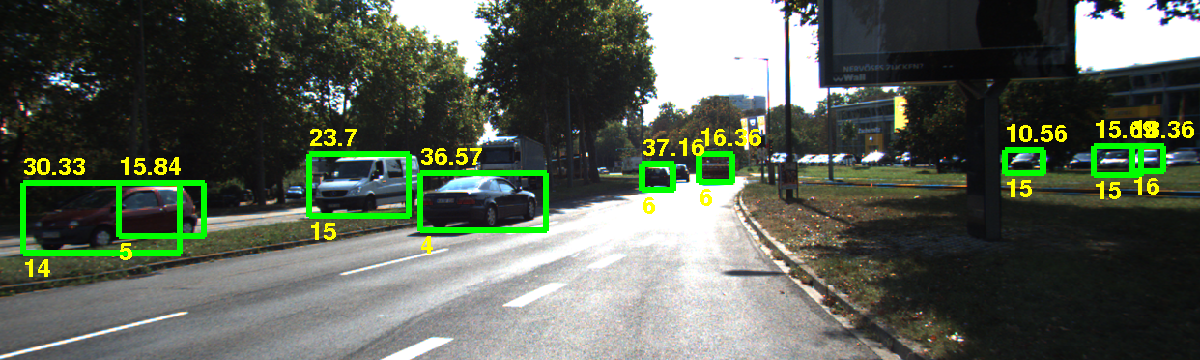}\\
  \vspace{0.4ex}
  \includegraphics[width=0.49\columnwidth]{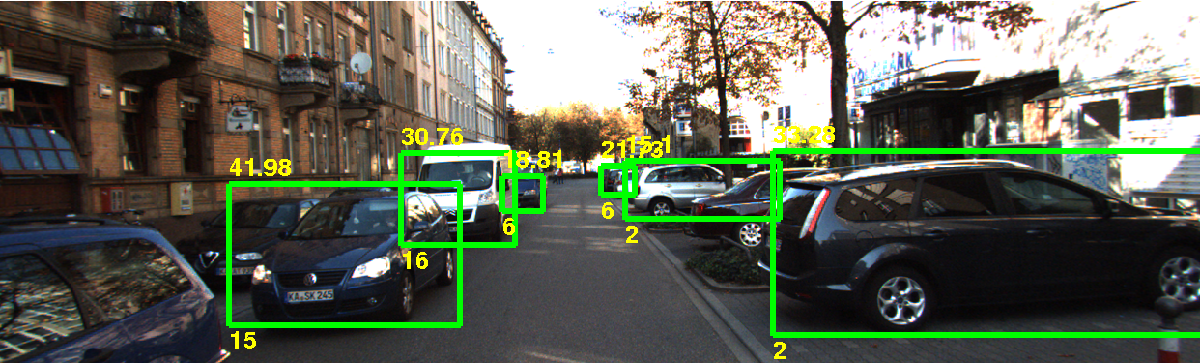}  
  \includegraphics[width=0.49\columnwidth]{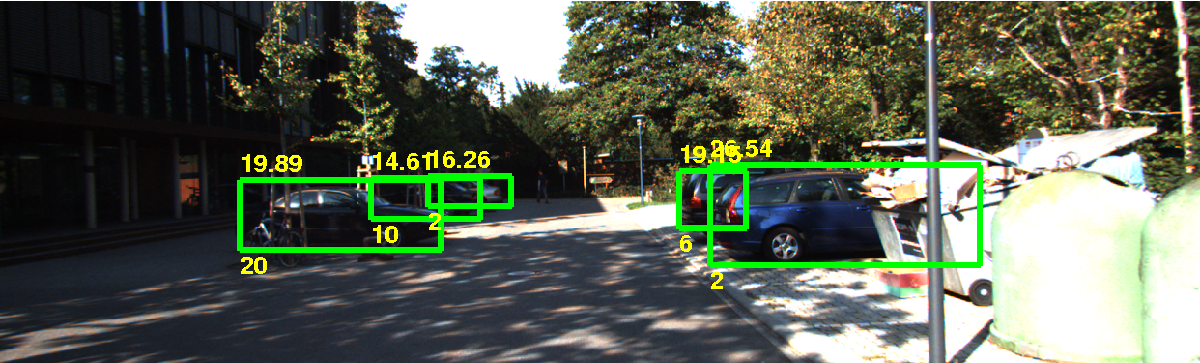}
  \caption{Qualitative detection results in
    KITTI-train90/KITTI-train10 experiment (shown detections with
    score$\geq$10).  For each detection, we show the score (above) and
    the estimated view class (below).}
  \label{fig:KITTI_TRAINING_QUALITATIVE}
\end{figure}

In Fig.~\ref{fig:KITTI_BADACOST_TREE_DEPTH} we can see an improvement
in the AP with deeper trees for BAdaCost 1-3-3. There is no
improvement in the recall from $D$=7 to $D$=9. However, in terms of
precision, there is a big improvement from a recall value of 0.6 and
above. On the other hand, there is no reason to prefer depth $D=9$ to
$D=8$ since both plots are almost equal.  Thus we can conclude that
the best BAdaCost classifier is the one with
$\alpha-\beta-\gamma$=1-3-3, $T=1024$, $D=8$ with an AP of 82.1 in the
KITTI-train90/KITTI-train10 experiments (see qualitative result in
Fig.~\ref{fig:KITTI_TRAINING_QUALITATIVE}).

To make a fair comparison, using the same kind of features and search
window size, we have trained a \emph{SubCat} detector with 20 car views (car
images 48 pixels heigh).  We use AdaBoost with 2048 maximum number of
decision trees each. In Fig.~\ref{fig:KITTI_SUBCAT} we can see the
results in the validation subset with depth $D$=2 and $D$=3. Finally,
we show the results of the best SubCat, SAMME and BAdaCost
configurations in Fig.~\ref{fig:KITTI_SAMME_VS_BADACOST_VS_SUBCAT} for
the moderate samples and in Table~\ref{tab:RESULTS_KITTI} for all
KITTI experiments. The \texttt{train90/train10} data set represents the
results of the experiments here described, while \texttt{train90/testing}
the results produced by the evaluation 
server~\footnote{\url{http://www.cvlibs.net/datasets/kitti/eval\_object.php} (\emph{SubCat48LDCF} and 
\emph{BdCost48LDCF}).}.

We can conclude that the detector built with a
cost-less multi-class AdaBoost, SAMME, is not better than the one built
with a set of binary detectors, SubCat. However, the introduction of
costs with BAdaCost produces a improvement of about 7 points in AP.
This experiment talks for itself about the benefits of using
multi-class cost-sensitive Boosting algorithms in asymmetric computer
vision problems.

\begin{table}[t]
\renewcommand{\arraystretch}{0.8}
\footnotesize
\centering
\begin{tabular}{c|c|ccc}
  {\bf Data sets}                     & {\bf Algorithm} & {\bf Easy} & {\bf Moderate} & {\bf Hard} \\
  \hline 
  \multirow{3}{*}{train90/train10}   & \small SubCat   & 80.00 \%      & 74.50 \%      & 58.30 \%\\
                                     & \small SAMME    & 81.00 \%      & 75.40 \%      & 58.80 \%\\
                                     & \small BAdaCost & {\bf84.8 \%} & {\bf82.10 \%} & {\bf66.90 \%}\\
  \hline 
  \multirow{2}{*}{train90/testing}   & \small SubCat   & 68.71 \%      & 61.79 \%      & 47.46 \%\\
                                     & \small BAdaCost & {\bf77.37 \%} & {\bf66.66 \%} & {\bf55.51 \%} \\
\end{tabular}
\caption{AP for the experiments on the KITTI data set.}
\label{tab:RESULTS_KITTI}
\end{table}


\subsubsection{Classification efficiency}
\label{Subsec:Efficiency}

Computational efficiency is a key issue in classification algorithms,
specially those used for object detection in a sliding window
approach. The computational cost of Boosting algorithm is directly
related to the number of decision tree nodes that have to be
evaluated to make a decision. 

In Table~\ref{Table:efficiency} we show the number of detectors,
\texttt{\#Detect}, number of tree weak-learners in each detector
\texttt{\#T}, tree depths \texttt{D} and number of node evaluations,
\texttt{\#Nodes} (number of weak-learners times tree depth times
number of detectors).

As shown in Table~\ref{Table:efficiency}, in the worst case, BAdaCost
evaluates fewer decision tree nodes than any of the
algorithms discussed in this section.
In the face detection problem, 
it is 3.25 more efficient than SquaresChnFtrs-5, 6.66 times more
efficient than SubCat and 14.32 times more efficient than
HeadHunter. In the car detection problem, it is 4.75 times more
efficient than our trained SubCat detector with LDCF features.

The multi-class cost-sensitive approach presented in this paper
is not only more flexible than its binary competitors learning 
class boundaries, but it also is far more efficient. We can
train BAdaCost in roughly one fourth of the time required for
SubCat.

\begin{table}[t]
\renewcommand{\arraystretch}{0.8}
\footnotesize
\centering
\begin{tabular}{c|c|cccc}
  \textbf{Data sets}   & \textbf{Algorithm} & \texttt{\#Detect} & \texttt{\#T} & \texttt{D} &\texttt{\#Nodes} \\
  \hline 
  \multirow{2}{*}{AFLW}              & SqChnnFtrs-5    & 5                & 2000     & 2        & 20000\\
                                     & SubCat          & 5                & 4096     & 2        & 40960\\
                                     & HeadHunter      & 22               & 2000     & 2        & 88000\\
				     & BAdaCost        & 1                & 1024     & 6        & {\bf 6144}\\
  \hline 
  \multirow{2}{*}{KITTI train90}     & SubCat          & 20               & $\leq$2048     & 2  & 38984\\
                                     & BAdaCost        & 1                & 1024     & 8        & {\bf 8192}\\
\end{tabular}
\caption{Number of decision tree nodes executed to classify.}
\label{Table:efficiency}
\end{table}

\section{Conclusions}
\label{Sec:Conclusions}

In this paper we have addressed the problem of extending the notion of
multi-class margin to the cost-sensitive margin. We have introduced
the BAdaCost algorithm and we have explored its theoretical
connections proving that it generalizes SAMME~\cite{Zhu09},
Cost-sensitive AdaBoost~\cite{Masnadi11} and PIBoost~\cite{Baldera14}.

The cost-sensitive multi-class approach introduced in this paper fills
an existing gap in the literature. It provides a sound Boosting
algorithm for solving multi-class cost-sensitive problems such as the
multi-view object detection.  The importance of a cost-sensitive
multi-class approaches comes from the fact that the cost matrix can be
used as a tool to learn class boundaries. There are relevant
multi-class problems for which this is important: class imbalance
problems~\cite{Baldera2015}, asymmetric classification like in 
detection and segmentation or any problem in which the
objective is not the global error but a metric like Jaccard or
F1-score~\cite{Parambath2014}.

We have shown experimentally that BAdaCost outperforms other relevant
multi-class Boosting algorithms in the literature:
Ada.C2M1~\cite{Sun06}, MultiBoost~\cite{Wang13} and
$L_p$-CSB~\cite{Lozano08}.  We have tested our cost-sensitive approach
in face and car detection problems.  The usual approach to multi-view
object detection is to use a binary detector per view. In our
experiments we have shown that for face and car detection, the
multi-class cost-sensitive Boosting improves by a large margin the
usual multi-view approach. This is a relevant outcome since one
limitation of current cascade architectures is the difficulty of
implementing multi-class detectors~\cite{Cai2016}.  Furthermore, our
algorithm uses much fewer weak-learners than the usual approaches to
multi-view object detection.


\section*{Acknowledgment}

The authors thank the anonymous reviewers for their positive comments.
This research was funded by the spanish \emph{Ministerio de
  Econom\'{\i}a y Competitividad}, projects 
\texttt{TIN2013-47630-C2-2-R} and
\texttt{TIN2016-75982-C2-2-R}. Jos\'e M.  Buenaposada acknowledges the
support of Computer Vision and Image Processing research group (CVIP)
from Universidad Rey Juan Carlos.

\bibliographystyle{elsarticle-num}
\bibliography{paper}

\appendix

\section{Proof of BAdaCost Lemma}
\label{Proof:Lemma}

Let us assume known $\vf_m(\vx)$, the stage-wise additive model up to
iteration $m$. We get
$\vf_{m+1}(\vx) = \vf_m (\vx) + \beta \vg (\vx)$, where $(\beta, \vg)$
minimize the expression
{\footnotesize
\[
\begin{split}
&\sum^N_{n=1} \exp\left( \vC^*(l_n,-)\left(\vf_m (\vx_n) + \beta \vg (\vx_n) \right) \right) \\
&= \sum^N_{n=1} w(n) \exp\left( \beta \vC^* (l_n,-)\vg (\vx_n) \right) \\
&= \sum^K_{j=1} \sum_{ \left\{ n:l_n = j \right\} } w(n) \exp\left( \beta \vC^* (j,-)\vg(\vx_n) \right) \\
&= \sum^K_{j=1} \sum_{ \left\{ n:l_n = j \right\} } w(n) \exp\left( \frac{\beta K}{K-1} C^* (j,G(\vx_n)) \right),
\end{split}
\]}
where $w(n)=\exp(\vC^*(l_n,-)\vf_m (\vx_n))$.

In the last step of the previous expression we take into account the
equivalence between vector-valued functions, $\vg$, and
label-valued functions, $G$.  Let
$\mathit{S}(j) = \left\{ n : l_n = G(\vx_n) = j \right\} $ be the set
of indexes of well classified instances with $l_n = j$ and let
$\mathit{F}(j,k) = \left\{ n : l_n = j, G(\vx_n) = k \right\}$ be the
indexes where $G$ outputs $k$ when the real label is $j$. Therefore,
we can rewrite the above expression as
{\footnotesize
\[
\begin{split}
&\sum^K_{j=1} \sum_{ n \in \mathit{S}(j)  } w(n) \exp\left( \frac{\beta K C^* (j,j) }{K-1}  \right) \\
&+ \sum^K_{j=1} \sum_{ n \in \mathit{F}(j,k) } w(n) \exp\left( \frac{\beta K C^* (j,k) }{K-1} \right) \\
&= \sum^K_{j=1} S_j \exp \left( \frac{\beta K C^* (j,j) }{K-1} \right) \\
&+ \sum^K_{j=1}\sum_{k \neq j} E_{j,k} \exp \left( \frac{\beta K C^* (j,k) }{K-1} \right).
\end{split}
\]}
Where $S_j = \sum_{ n \in \mathit{S}(j) } w(n)$ and $E_{j,k} = \sum_{ n \in
\mathit{F}(j,k) } w(n)$.
Taking into account that these constants are positive values and also
$\exp(\beta C^*(i,j)) > 0$  ($\forall i,j \in L$), we can omit the term
$K/(K-1)$ in the exponents to solve the minimization.
Subsequently, the objective function can be written:
\begin{equation} \label{Exp2}
\sum^K_{j=1} \left( S_j \exp \left(\beta C^* (j,j) \right) + \sum_{k \neq j} E_{j,k} \exp \left(\beta C^* (j,k) \right) \right).
\end{equation}
Now, fixed a value $\beta > 0$, the optimal step, $\vg$, can be found minimizing
{\footnotesize
\[
\begin{split}
&\sum^K_{j=1} \left( S_j \underbrace{ \exp \left( \beta C^*(j,j) \right) }_{A(j,j)^{\beta}} + \sum_{k \neq j} E_{j,k} \underbrace{ \exp \left( \beta C^*(j,k) \right) }_{A(j,k)^\beta}  \right) \\
&= \sum^K_{j=1} \left( \left( \sum_{ n \in \mathit{S}(j) } w(n) \right)  A(j,j)^{\beta} + \sum_{k \neq j} \left( \sum_{ n \in \mathit{F}(j,k) } w(n) \right)  A(j,k)^\beta  \right) \\
&= \sum^N_{n=1} w(n) \left( A(l_n,l_n)^{\beta} I \left( G(\vx_n) = l_n \right) + \sum_{k \neq l_n} A(l_n,k)^\beta I\left( G(\vx_n) = k \right)  \right).
\end{split}
\]}
Finally if we assume known a direction, $\vg$, then its weighted errors,
$E_{j,k}$, and successes, $S_j$, will be computable.
So deriving (\ref{Exp2}) with respect to $\beta$ (note that is a convex
function) and equating to zero we get
{\footnotesize
\[
\begin{split}
&\sum^K_{j=1} \left( \sum_{k \neq j} E_{j,k} C^*(j,k)\exp \left( \beta C^*(j,k) \right) + S_j C^*(j,j) \exp \left( \beta C^*(j,j) \right) \right)= 0 \\
&\sum^K_{j=1} \sum_{k \neq j} E_{j,k} C^*(j,k) \exp \left( \beta C^*(j,k) \right) = -\sum^K_{j=1} S_j C^*(j,j)\exp \left( \beta C^*(j,j) \right) \\
&\sum^K_{j=1} \sum_{k \neq j} E_{j,k} C^*(j,k) A(j,k)^\beta = -\sum^K_{j=1} S_j C^*(j,j) A(j,j)^{\beta} \\
&\sum^K_{j=1} \sum_{k \neq j} E_{j,k} C(j,k) A(j,k)^\beta = \sum^K_{j=1} \sum^K_{h=1} S_j C(j,h) A(j,j)^{\beta},
\end{split}
\]}
where $A(j,k)=\exp C^*(j,k)$.


\section{Proof of Corollary 1 (SAMME is a special case of BAdaCost)}

It is easy to see that when $\vC$ is defined in the following way:
\begin{equation}
C(i,j) := \left\{
  \begin{array}{l l}
   0 & \quad \mbox{for} \: i = j\\
   \frac{1}{K(K-1)} & \quad \mbox{for} \: i \neq j\\
  \end{array} \right.
\; \; \forall i,j \in L,
\end{equation}
then a discrete vectorial weak learner, $\vf$, yields $\vC^* (l,-) \vf (\vx) =
-1/(K-1)$ for correct classifications and $\vC^* (l,-) \vf(\vx) = 1/(K-1)^2$ for
errors. Both quantities are, in fact, the possible values of $\frac{-1}{K}\vy^\top
\vf(\vx)$ in the exponent of the loss function in SAMME.
Thus expression (\ref{Exp2}) can be written
{\footnotesize
\[
\begin{split}
&\sum^{K}_{j=1} \left( S_j\exp \left( \frac{-\beta}{K-1} \right) + \sum_{k \neq j} E_{j,k} \exp \left( \frac{\beta}{(K-1)^2} \right)  \right) \\
&= \underbrace{ \left( \sum^{K}_{j=1} S_j \right) }_{S} \exp \left( \frac{-\beta}{K-1} \right) +  \underbrace{ \left( \sum^{K}_{j=1} \sum_{k \neq j} E_{j,k} \right) }_{E} \exp \left( \frac{\beta}{(K-1)^2} \right) \\
&= S \exp \left( \frac{-\beta}{K-1} \right) +  E \exp \left( \frac{\beta}{(K-1)^2} \right) \\
&= (1 - E) \exp \left( \frac{-\beta}{K-1} \right) +  E \exp \left( \frac{\beta}{(K-1)^2} \right) \\
&= \exp \left( \frac{-\beta}{K-1} \right) + E \left( \exp \left( \frac{\beta}{(K-1)^2} \right) - \exp \left( \frac{-\beta}{K-1} \right) \right) .
\end{split}
\]}
So, the above expression is minimized when
$E = \sum^N_{n = 1} w(n) I \left( G(\vx_n) \neq l_n \right)$ is
minimum.  For the second point of the corollary we just need to
consider the above expression as a function of $\beta$.
Computing the derivative and setting equal to zero we get
{\footnotesize
\[
\begin{split}
&\frac{E}{K-1}\exp \left( \frac{\beta}{(K-1)^2} \right) = (1 - E) \exp \left( \frac{-\beta}{K-1} \right) \\
&\exp \left( \frac{K \beta}{(K-1)^2} \right) = \frac{(K - 1)(1 - E)}{E},
\end{split}
\]}
and taking logarithms
{\footnotesize
\[
\begin{split}
&\frac{K \beta}{(K-1)^2} = \log \left( \frac{1 - E}{E} \right) + \log \left( K - 1 \right) \\
&\beta = \frac{(K-1)^2}{K} \left( \log \left( \frac{1 - E}{E} \right) + \log \left( K - 1 \right) \right).
\end{split}
\]}
Hence the second point of the corollary follows.


\section{Proof of Corollary 2 (CS-AdaBoost is a special case of BAdaCost)}

Given the ($2 \times 2$)-cost-matrix, let $C_1 = C(1,2)$ and $C_2 = C(2,1)$ denote the non diagonal values.
The expression (\ref{Exp2}) becomes:
{\footnotesize
\[
\begin{split}
&\sum^{2}_{j=1} \left( S_j\exp \left( -\beta C_j \right) + \underbrace{E_{j,k} }_{k \neq j} \exp \left( \beta C_j \right)  \right) = \\
&S_{1}\exp \left( -\beta C_1 \right) + E_{1,2} \exp \left( \beta C_1 \right) + S_{2}\exp \left( -\beta C_2 \right) + E_{2,1} \exp \left( \beta C_2 \right) .
\end{split}
\]}
Let us assume now that $\beta > 0$ is known. Using Badacost Lemma, the
optimal discrete weak learner minimizing the expected loss is
{\footnotesize
\[
\arg\min_{\vg} \left[ \mathrm{e}^{ \beta C_1 } E_{1,2} + \mathrm{e}^{ -\beta C_1 } S_1 + \mathrm{e}^{ \beta C_2 } E_{2,1} + \mathrm{e}^{ -\beta C_2 } S_2 \right],
\]}
changing the notation $T_1 = \sum_{ \{ n : l_n = 1 \} } w(n)$, $T_2 = \sum_{ \{ n : l_n = 2 \} } w(n)$, $E_{1,2} = b = T_1 - S_1$ and $E_{2,1} = d = T_2 - S_2$,
{\footnotesize
\[
\arg\min_{\vg} \left[ \mathrm{e}^{ \beta C_1 } b + \mathrm{e}^{ -\beta C_1 } (T_1 - b) + \mathrm{e}^{ \beta C_2 } d + \mathrm{e}^{ -\beta C_2 } (T_2 - d)\right] =
\]}
{\footnotesize
\begin{equation} \label{Equa:G_optimaBina}
\begin{split}
\arg\min_{\vg}~[
 \left( \mathrm{e}^{ \beta C_1 } - \mathrm{e}^{ -\beta C_1 } \right) b ~~~~~~~~~~~~~~~~~~~~~~~~~~~~~~~~~~~~~~~~~~~~~~~~~~~~~~~ \\
               + \mathrm{e}^{ -\beta C_1 } T_1 + \left( \mathrm{e}^{ \beta C_2 }
               - \mathrm{e}^{-\beta C_2 } \right) d + \mathrm{e}^{ -\beta C_2 } T_2
].
\end{split}
\end{equation}}
Besides, if we assume known the optimal weak learner, $\vg$, then its
weighted success/error rates will be computable. We can find the best
value $\beta$ using BAdaCost Lemma. In this binary case the following
expression must be solved
{\footnotesize
\[
E_{1,2}C_1 \mathrm{e}^{ \beta C_1 } + E_{2,1}C_2 \mathrm{e}^{ \beta C_2 } = S_1 C_1 \mathrm{e}^{ -\beta C_1 } + S_2 C_2 \mathrm{e}^{ -\beta C_2 } .
\]}
Again, using the notation $T_1$, $T_2$, $b$ and $d$ we get
{\footnotesize
\[
\begin{split}
&b C_1 \mathrm{e}^{ \beta C_1 } + d C_2 \mathrm{e}^{ \beta C_2 } = (T_1 - b) C_1 \mathrm{e}^{ -\beta C_1 } + (T_2 - d) C_2 \mathrm{e}^{ -\beta C_2 } \\
&b C_1 \left( \mathrm{e}^{ \beta C_1 } + \mathrm{e}^{ -\beta C_1 } \right) + d C_2 \left( \mathrm{e}^{ \beta C_2 } + \mathrm{e}^{ \beta C_2 } \right) = T_1 C_1 \mathrm{e}^{ -\beta C_1 } + T_2 C_2 \mathrm{e}^{ -\beta C_2 }
\end{split}
\]}
{\footnotesize
\begin{equation} \label{Equa:Beta_optimaBina}
2 b C_1 \cosh \left( \beta C_1 \right) + 2 d C_2 \cosh \left( \beta C_2 \right) =  T_1 C_1 \mathrm{e}^{ -\beta C_1 } + T_2 C_2 \mathrm{e}^{ -\beta C_2 }\:,
\end{equation}}%
that proves the equivalence between both algorithms for binary problems.


\section{Proof of Corollary 3 (PIBoost is a special case of BAdaCost)}
\label{Proof:Coro3}

Let $S$ denote a subset of $s$-labels of the problem.
The margin values are $\vy^\top \vf(\vx) = \frac{\pm K}{s(K-1)}$, when $\vy \in
S$, and $\vy^\top \vf(\vx) = \frac{\pm K}{(K-s)(K-1)}$, when $\vy \notin S$. In
both cases there is a positive/negative sign in case of correct/wrong
classification.
In turn the exponential loss function, $\exp( -\vy^\top \vf(\vx)/K)$, yields
$\exp ( \frac{\mp 1}{s(K-1)} )$ and $\exp ( \frac{\mp 1}{(K-s)(K-1)} )$
respectively.
Let $\vC$ be the $(2 \times 2)$-cost-matrix with non diagonal values $C(1,2) =
\frac{1}{sK}$ and $C(2,1) = \frac{1}{(K-s)K}$.
This matrix produces cost-sensitive multi-class margins with the same values on
the loss function.
Thus we can apply the Lemma to this binary cost-sensitive sub-problem. In
particular we can apply Corollary 2 directly.
Replacing in expression (\ref{Equa:G_optimaBina}) we get the optimal weak
learner, $\vg$, solving
{\footnotesize
\begin{equation}
\begin{split}
&\arg\min_{\vg} \left( \mathrm{e}^{\frac{\beta}{s(K-1)}} - \mathrm{e}^{\frac{-\beta}{s(K-1)}} \right) E_1 + A_1 \mathrm{e}^{\frac{-\beta}{s(K-1)}} + \\
&+ \left( \mathrm{e}^{\frac{\beta}{(K-s)(K-1)}} - \mathrm{e}^{\frac{-\beta}{(K-s)(K-1)}} \right) E_2 + A_2 \mathrm{e}^{\frac{-\beta}{(K-s)(K-1)}} \: .
\end{split}
\end{equation}}
Where $A_1 = \sum_{ \{ n : l_n = 1 \} } w(n)$, $A_2 = \sum_{ \{ n : l_n = 2 \} }
w(n)$, $E_1 = \sum_{ \{ n : g(x_n) \neq l_n = 1 \} } w(n)$ and $E_2 = \sum_{ \{
n : g(x_n) \neq l_n = 2 \} } w(n)$.
If we assume known the optimal direction of classification $\vg$, then its
weighted errors and successes, we can compute the optimal step $\beta$ using
(\ref{Equa:Beta_optimaBina}) as the solution to
{\footnotesize
\[
\begin{split}
&\frac{2E_1}{s} \cosh \left( \frac{\beta}{s(K-1)} \right) + \frac{2E_2}{(K-s)} \cosh \left( \frac{\beta}{(K-s)(K-1)} \right) = \\
&\frac{A_1}{s} \exp\left( \frac{-\beta}{s(K-1)} \right) + \frac{A_2}{(K-s)} \exp\left( \frac{-\beta}{(K-s)(K-1)} \right) \: .
\end{split}
\]}
%
Denoting $\beta = s(K-s)(K-1) \log x$ we get
{\footnotesize
\[
\frac{E_1}{s} \left(  x^{(K-s)} - x^{-(K-s)} \right) + \frac{E_2}{(K-s)} \left( x^s - x^{-s} \right) = \frac{A_1}{s} x^{-(K-s)} + \frac{A_2}{(K-s)} x^{-s}\: .
\]}
Which is equivalent to finding the only real solution (Descartes Theorem of signs)
of the following polynomial:
{\footnotesize
\[
P(x) = E_1 (K-s)x^{2(K-s)} + E_2 s x^K - s(A_2 - E_2)x^{(K-2s)} - (K-s)(A_1 - E_1) \; .
\]}
Hence the Corollary follows.

\section{UCI Data Bases}

In Section~5.1 of the paper we we select 12 data sets from the UCI
repository that cover a broad range
of multi-class classification problems with regard to the number of
variables, labels, and data instances. In Table~\ref{table:bench_Cost}
we provide additional information about them.

\begin{table}[ht]
\renewcommand{\arraystretch}{0.8}
\footnotesize
\centering
\begin{tabular}{l|ccc}
{\bf Data set} &{\bf Variables} & {\bf Labels} & {\bf Instances} \\ \hline
CarEvaluation  & 6     & 4  & 1728  \\
Chess          & 6     & 18 & 28056 \\
CNAE9          & 856   & 9  & 1080  \\
ContraMethod   & 9     & 3  & 1473  \\
Isolet         & 617   & 26 & 7797  \\
Letter         & 16    & 26 & 20000 \\
Shuttle        & 9     & 7  & 58000 \\
OptDigits      & 64    & 10 & 5620  \\
PenDigits      & 16    & 10 & 10992 \\
SatImage       & 36    & 7  & 6435  \\
Segmentation   & 19    & 7  & 2310  \\
Waveform       & 21    & 3  & 5000  \\
\end{tabular}
\caption{Summary of selected UCI data sets}
\label{table:bench_Cost}
\end{table}

\section{A cost matrix for imbalanced classification problems\protect\footnote{From Section~5 of~\cite{Baldera2015}}}

A preliminary issue when using a cost-sensitive algorithm for solving
an imbalance problem is establishing the cost matrix, $\vC$.  A
straightforward solution would be to set the costs inversely
proportional to the class imbalance ratios. However, this solution
does not take into account the complexity of the classification
problem.  i.e.  the amount of class overlap, within-class imbalance,
etc.  Here we introduce an alternative solution that considers the
problem complexity. To this end we introduce a cost matrix that weighs
more heavily the errors of poorly classified classes, hence the
classifier will concentrate on the difficult minority classes.

Let $\vF$ be the confusion matrix and $\vF^*$ the matrix obtained when
dividing each row $i$, $\vF(i,-)$, by $\vF(i,\cdot)=\sum_jF(i,j)$,
i.e. the number of samples in class $i$. Then $F^*(i,j)$ is the
proportion of data in class $i$ classified as $j$. In a complex and
imbalanced data-set, a $0|1$-loss classifier (e.g. BAdaCost with
$0|1$-losses) will tend to over-fit the majority classes. So,
off-diagonal elements in rows $\vF^*(i,-)$ for majority
(alt. minority) classes will have low (high) scores. Hence, the
resulting matrix after setting $F^*(i,i)=0, \forall i=1\ldots K$ is
already a cost matrix.  Finally, to improve numerical conditioning, we
set $\vC=\lambda\vF^*$, for a small $\lambda>0$.

\section{Face detector trained with BAdaCost}

Classifiers trained with BAdaCost have a number of interesting
features to explore. We first show additional information about the
face detector learned with the AFLW database and used in the
experiments in the paper. Afterwards we also include information about
the car detector trained with KITTI-train90, also used in the experiments.

\begin{figure}[htbp]
  \centering
  \includegraphics[width=0.3\textwidth]{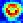}
  \includegraphics[width=0.3\textwidth]{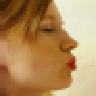}
  \includegraphics[width=0.3\textwidth]{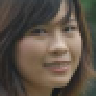} \\
  \vspace{0.5ex}
  \includegraphics[width=0.3\textwidth]{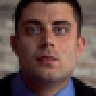}
  \includegraphics[width=0.3\textwidth]{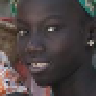}
  \includegraphics[width=0.3\textwidth]{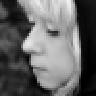}
  \caption{Spatial distribution of features selected by BAdaCost in
    AFLW (top left).  Red color represent the most frequently used
    features, blue less frequently used ones.  We also show several
    face images to compare them with the feature location map.}
  \label{fig:AFLW_FeatMap_Full}
\end{figure}

\begin{figure}[htbp]
  \centering
  \setlength\tabcolsep{0.5ex} 
  \begin{tabular}{cccccccccc}
  \includegraphics[width=0.19\textwidth]{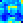} &
  \includegraphics[width=0.19\textwidth]{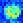} &
  \includegraphics[width=0.19\textwidth]{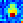} &
  \includegraphics[width=0.19\textwidth]{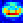} &
  \includegraphics[width=0.19\textwidth]{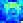} \\
    L & U & V & $||.||$ & $90^o$\\
  \includegraphics[width=0.19\textwidth]{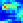} &
  \includegraphics[width=0.19\textwidth]{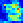} &
  \includegraphics[width=0.19\textwidth]{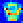} &
  \includegraphics[width=0.19\textwidth]{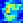} &
  \includegraphics[width=0.19\textwidth]{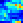} \\
   $60^o$ & $30^o$ & $0^o$ & $-30^o$ & $-60^o$
  \end{tabular}
  \caption{Per channel spatial distribution of features selected by
    BAdaCost in AFLW.  Red color represent the most frequently used
    features, blue less frequently used ones.}
  \label{fig:AFLW_PerFeatMap}
\end{figure}

Fig.~\ref{fig:AFLW_FeatMap_Full} (top left) shows the spatial
distribution of features selected by BAdaCost in the face detection
experiment. We use the weights of the weak learner to compute the
map. This means that a feature used in a weak learner tree contributes
to the map with the weak learner weight.  The more reddish is the color
of the corresponding spatial location in the map, the more times a
feature in this location has been selected by a weak learner with high
weight.  On the other hand, Fig.~\ref{fig:AFLW_PerFeatMap} shows the
spatial distribution of face features selected in each of the original
ACF channels: color (LUV), gradient magnitude ($||.||$) and gradient
filters at different orientations. In Fig.~\ref{fig:AFLW_PerFeatMap}
we can see that the color channels are symmetrical in terms of feature
localization, that is reasonable since skin color is independent of face
orientation. The magnitude of the gradient in the center of the face
is also important for detection: the face has strong edges
around the mouth and nose. Finally, the importance of gradient
orientation channels depends on the face orientation class
(e.g. $30^o$ edges are important for half profile faces while $0^o$
edges are important for frontal faces.).

\begin{figure}[htbp]
  \centering
    \includegraphics[width=0.44\textwidth]{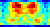}
    \includegraphics[width=0.43\textwidth]{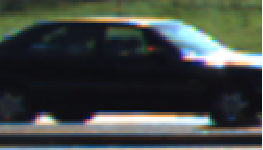} \\
  \vspace{0.5ex}
    \includegraphics[width=0.43\textwidth]{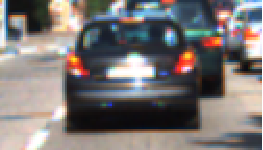}
    \includegraphics[width=0.43\textwidth]{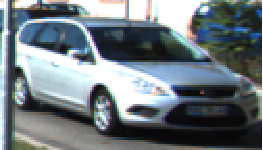}
    \caption{Spatial distribution of selected features by the BAdaCost
      algorithm in KITTI-train90 (top left).  Red color represent the most
      frequently used features, blue less frequently used ones.  We
      also show training face images to compare them with the feature
      location map.}
  \label{fig:KITTI_FeatMap_Full}
\end{figure}

\begin{figure}[htbp]
  \centering
  \setlength\tabcolsep{0.5ex} 
  \begin{tabular}{ccccc}
    \includegraphics[width=0.15\textwidth]{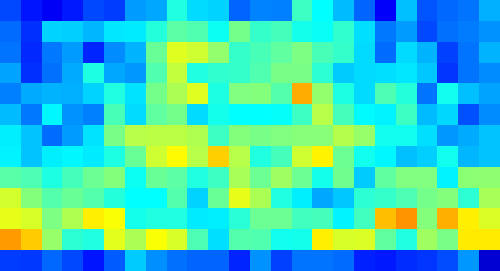} &
    \includegraphics[width=0.15\textwidth]{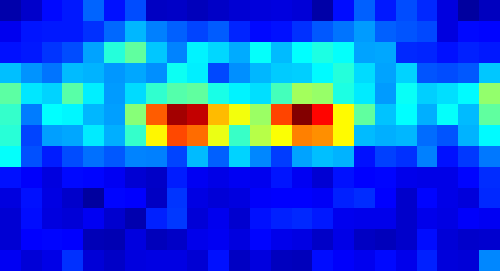} &
    \includegraphics[width=0.15\textwidth]{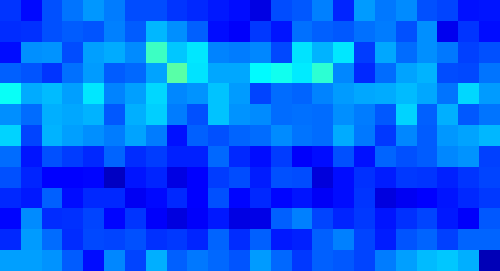} &
    \includegraphics[width=0.15\textwidth]{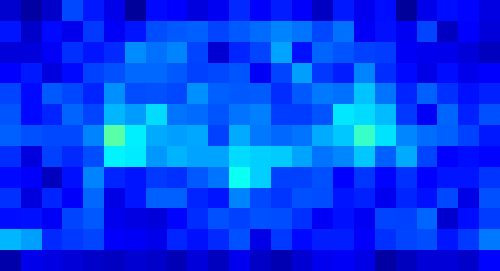} &
    \includegraphics[width=0.15\textwidth]{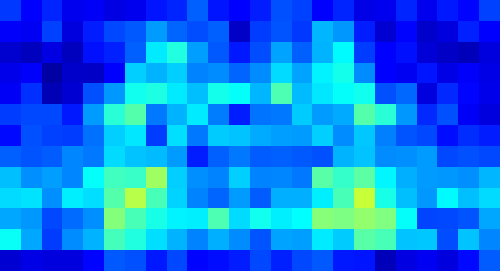} \\
     L & U & V & $||.||$ & $950^o$ \\
  \includegraphics[width=0.15\textwidth]{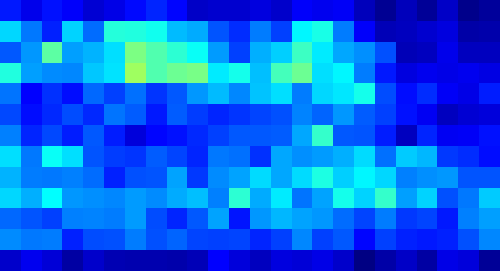} &
  \includegraphics[width=0.15\textwidth]{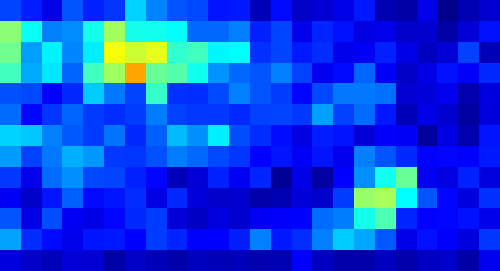} &
  \includegraphics[width=0.15\textwidth]{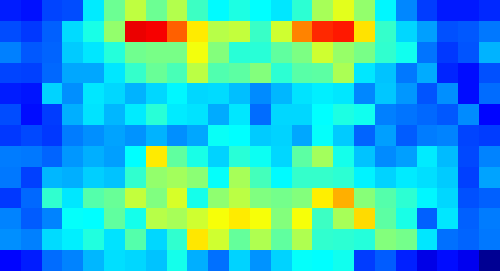} &
  \includegraphics[width=0.15\textwidth]{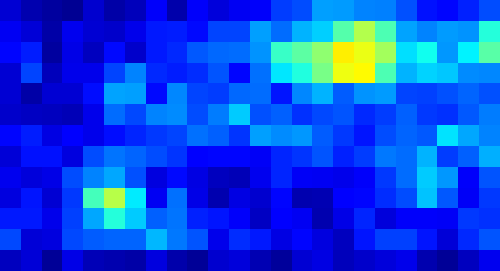} &
  \includegraphics[width=0.15\textwidth]{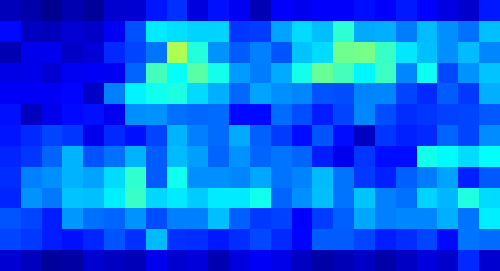} \\
   $60^o$ & $30^o$ & $0^o$ & $-30^o$ & $-60^o$
  \end{tabular}
  \caption{Per channel spatial distribution of selected features by
    the BAdaCost algorithm in KITTI-train90.  Red color represent the
    most frequently used features, blue less frequently used ones.}
  \label{fig:KITTI_PerFeatMap}
\end{figure}

Fig.~\ref{fig:KITTI_FeatMap_Full} shows the spatial distribution of
car features selected by BAdaCost. Fig.~\ref{fig:KITTI_PerFeatMap}
shows the spatial distribution of features selected in each of the
original ACF channels. In Fig.~\ref{fig:KITTI_PerFeatMap} we can see
that color channels have a global role in car detection (see L
channel).  However, the U channel seems to be used to detect the car
rear lights.  Also, as in the face detector, gradient orientation
channels respond selectively to the car orientation.  For example, the
$30^o$ channel responds to view 17 or 18 (half right profile).

\section{Cascade calibration}

\begin{figure}[htbp]
  \centering
   \includegraphics[width=0.8\columnwidth]{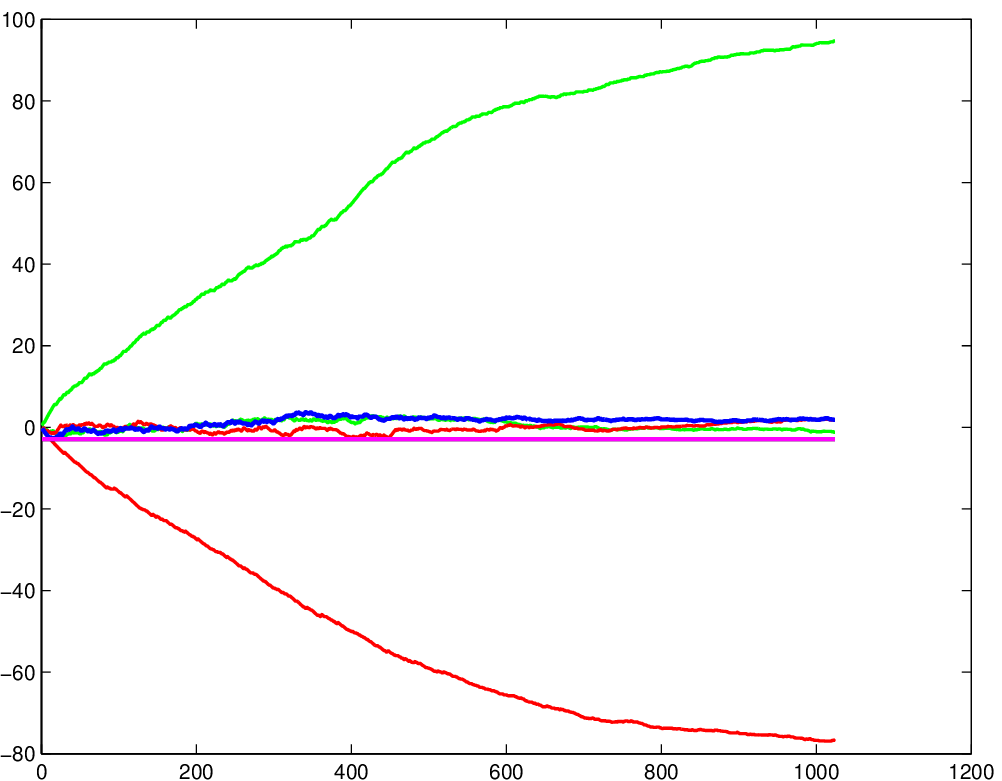}
   \caption{Score as a function of weak learner evaluated
     (i.e. example trace) for the best BAdaCost classifier trained
     with the KITTI-train90 database.}
  \label{fig:KITTI_Learning}
\end{figure}

Boosting classifiers used in detection can execute all weak learners
or stop when we are sure that the image window is negative. In
Fig.~\ref{fig:KITTI_Learning} we show the score in the $x$ axis and
the number of executed weak learners in the $y$ axis. In it we plot
the largest and lowest score for positive examples (in green) and the
lowest and largest score for negatives (in red). It is clear that
negative examples scores have lower values than those of the positive
ones. So, we can stop evaluating weak learners when the score falls
below a calibrated threshold.

We use a slightly modified version of Direct Backward Pruning
(DBP)~\cite{Zhang07}. The DBP algorithm
finds the positive training example with
the lowest score in the last weak learner (LPSE, Lowest Positive Score
Example), see blue plot in Fig.~\ref{fig:KITTI_Learning}.
In our case, we set a threshold that corresponds to the lowest LPSE (see purple horizontal line in
Fig.~\ref{fig:KITTI_Learning}).

\end{document}

%% file: labvision_defs.tex
\def\mat#1{\mathchoice{\mbox{\boldmath $\displaystyle\tt#1$}}
{\mbox{\boldmath$\textstyle\tt#1$}}
{\mbox{\boldmath$\scriptstyle\tt#1$}}
{\mbox{\boldmath$\scriptscriptstyle\tt#1$}}}

\def\vect#1{\mathchoice{\mbox{\boldmath $\displaystyle\bf#1$}}
{\mbox{\boldmath  $\textstyle\bf#1$}}
{\mbox{\boldmath  $\scriptstyle\bf#1$}}
{\mbox{\boldmath  $\scriptscriptstyle\bf#1$}}}

\def\v0{{\vect 0}}
\def\vone{{\vect 1}}

\def\vc{{\vect c}}

\def\vf{{\vect f}}
\def\vg{{\vect g}}

\def\vw{{\vect w}}
\def\vx{{\vect x}}
\def\vy{{\vect y}}

\def\vA{{\vect A}}

\def\vC{{\vect C}}

\def\vF{{\vect F}}

\def\vP{{\vect P}}
\def\vM{{\vect M}}

\def\vX{{\vect X}}

\def\v0{{\vect 0}}

\def\vone{{\vect 1}}

\def\mDelta{{\mat\mDelta}}

\def\mP{{\mat P}}

\def\m1{{\mat 1}}

\def\cF{\mathcal{F}}

\def\cF{\mathcal{F}}

\def\cL{\mathcal{L}}

\def\cP{\mathcal{P}}

\def\cR{\mathcal{R}}